\documentclass[journal]{IEEEtran}

\usepackage{amsmath,amsfonts}
\usepackage{subfigure}
\usepackage{graphicx}
\usepackage{balance}
\usepackage[ruled,linesnumbered,boxed,noend]{algorithm2e}
\usepackage{booktabs}
\usepackage{threeparttable}
\usepackage{hyperref}
\usepackage{array}
\usepackage{textcomp}
\usepackage{stfloats}
\usepackage{url}
\usepackage{verbatim}
\usepackage{cite}
\usepackage{algpseudocode}
\usepackage{stfloats}

\begin{document}

\title{LVCP: LiDAR-Vision Tightly Coupled Collaborative Real-time Relative Positioning}

\author{Zhuozhu Jian$^{1}$, Qixuan Li$^{1}$, Shengtao Zheng$^{2}$, Xueqian Wang$^{1}$ and Xinlei Chen$^{3}$

\thanks{
This work was supported by the National Key R\&D program of China 2022YFC3300703, the Natural Science Foundation of China under Grant 62371269, Guangdong Innovative and Entrepreneurial Research Team Program 2021ZT09L197, Shenzhen 2022 Stabilization Support Program WDZC20220811103500001, and Tsinghua Shenzhen International Graduate School Cross-disciplinary Research and Innovation Fund Research Plan JC20220011.
\textit{(Zhuozhu Jian and Qixuan Li are co-first authors.) (Corresponding author: Xinlei Chen.)}}
\thanks{$^{1}$Zhuozhu Jian, Qixuan Li and Xueqian Wang are with the Shenzhen International Graduate School, Tsinghua University, Shenzhen 518055, China (e-mail: 
\href{mailto:jzz21@mails.tsinghua.edu.cn}{jzz21@mails.tsinghua.edu.cn}; \href{mailto:lqx23@mails.tsinghua.edu.cn}{lqx23@mails.tsinghua.edu.cn}; \href{mailto:wang.xq@sz.tsinghua.edu.cn}{wang.xq@sz.tsinghua.edu.cn}).}
\thanks{$^{2}$Shengtao Zheng is with the School of Mechanical Engineering and Automation at Harbin Institute of Technology, Shenzhen 518055, China (e-mail: \href{mailto:200320610@stu.hit.edu.cn}{200320610@stu.hit.edu.cn})). 
}
\thanks{$^{3}$Xinlei Chen is with the Shenzhen International Graduate School, Tsinghua University, Shenzhen 518055, China, Pengcheng Lab, Shenzhen 518055, China, RISC-V International Open Source Laboratory, Shenzhen 518055, China (e-mail: 
\href{mailto:chen.xinlei@sz.tsinghua.edu.cn}{chen.xinlei@sz.tsinghua.edu.cn}.}
}



\maketitle

\begin{abstract}
In air-ground collaboration scenarios without GPS and prior maps, the relative positioning of drones and unmanned ground vehicles (UGVs) has always been a challenge. For a drone equipped with monocular camera and an UGV equipped with LiDAR as an external sensor, we propose a robust and real-time relative pose estimation method (LVCP) based on the tight coupling of vision and LiDAR point cloud information, which does not require prior information such as maps or precise initial poses. Given that large-scale point clouds generated by 3D sensors has more accurate spatial geometric information than the feature point cloud generated by image, we utilize LiDAR point clouds to correct the drift in visual-inertial odometry (VIO) when the camera undergoes significant shaking or the IMU has a low signal-to-noise ratio. To achieve this, we propose a novel coarse-to-fine framework for LiDAR-vision collaborative localization. 
In this framework, we construct point-plane association based on spatial geometric information, and innovatively construct a  point-aided Bundle Adjustment (BA) problem as the backend to simultaneously estimate the relative pose of the camera and LiDAR and correct the VIO drift. In this process, we propose a particle swarm optimization (PSO) based sampling algorithm to complete the coarse estimation of the current camera-LiDAR pose. In this process, the initial pose of the camera used for sampling is obtained based on VIO propagation, and the valid feature-plane association number (VFPN) is used to trigger PSO-sampling process. Additionally, we propose a method that combines Structure from Motion (SFM) and multi-level sampling to initialize the algorithm, addressing the challenge of lacking initial values. We further extend this method to simultaneously perform multiple drones localization task. We validate the effectiveness, real-time performance, and robustness of the algorithm on both open-source datasets and self-built datasets. (Video and Datasets\footnote{Site: \url{https://sites.google.com/view/lvcp}.})
\end{abstract}

\begin{IEEEkeywords}
Air-ground Collaboration, Point-aided Optimization, LiDAR-Vision Relatve Positioning.
\end{IEEEkeywords}

\section{Introduction}
\label{sec:Introduction}

\IEEEPARstart{I}{n} recent years, there has been rapid advancement in unmanned aerial vehicle (UAV) localization and navigation technology. UAVs can efficiently survey extensive areas and provide advantageous perspectives. However, there are limitations such as short endurance and an inability to carry heavy payloads. Unmanned ground vehicles (UGVs) can effectively compensate for these limitations \cite{li2024collaborative}. Therefore, the collaboration between UAVs and UGVs has emerged as a key research focus. In many collaborative tasks, such as the takeoff, landing \cite{gao2023adaptive}, mapping \cite{michael2014collaborative}, and collaborative exploration \cite{li2023colag}\cite{wang2020collaborative} of UAVs, relative positioning requires higher accuracy than absolute positioning. So the relative real-time accurate positioning of UAVs and UGVs is an essential and fundamental problem and must be solved. 

\begin{figure}[t]
    \centering
    \includegraphics[width=9cm]{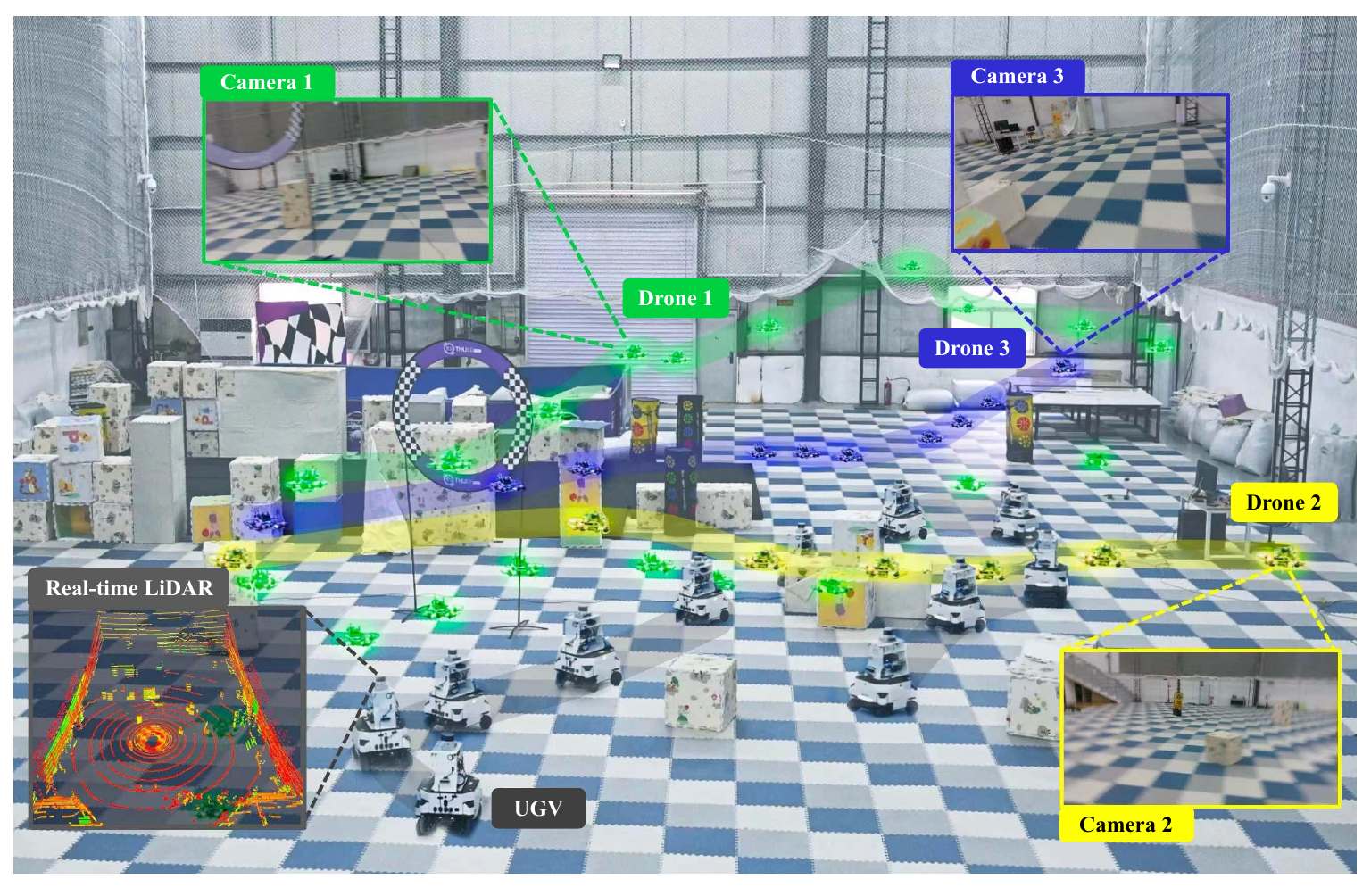}
    \caption{The LiDAR and vision-based collaborative positioning system (LVCP) in a UGV and multi-UAVs scenario. An accurate point cloud map is constructed in real-time using LiDAR mounted on a UGV. Multiple small UAVs collect images and IMU data, synchronously registering the images to the dynamic point cloud while the UGV is in motion, thus achieving relative localization among multiple agents. Experimental results validate that the LVCP system can achieve accurate relative localization of UAVs without prior map and initial pose disturbances.}
    \label{fig_exp_sen1}
    \vspace{-0.2cm}
\end{figure}

Due to the low cost and light weight, monocular cameras are widely used in UAVs, leading to the development of numerous monocular SLAM algorithms \cite{qin2018vins}\cite{campos2021orb}. However, monocular VIO systems cannot directly obtain absolute environmental information and cannot correct the estimation defects of monocular cameras and IMU. LiDAR is a commonly used external sensor for UGVs for its wide field of view and high-precision, high-resolution spatial information of the surrounding environment\cite{balestrieri2021sensors}. It can perceive the three-dimensional geometric structure of the surrounding environment in the form of point cloud, and can directly obtain high-precision scale information\cite{bai2023colmap}\cite{miao2024survey}. Therefore, in the relative pose estimation problem of UAV-UGV collaborative tasks, the absolute environmental information provided by LiDAR point clouds can be used to correct the scale estimation error and trajectory drift of monocular VIO algorithms. 

There exists two challenges with this approach: 1) In unknown scenarios, we cannot perform the priori processing on the point cloud (line feature extraction, cross-modal descriptors, etc.), which makes it difficult for monocular VIO systems to solve the drift problem with point clouds as the running time and trajectory length increase. 2) In real-world scenarios, the precise initial relative pose between the UAV and the UGV cannot be obtained accurately, leading to an accumulation of positioning inaccuracy over time. 


Some recent works focus on estimating the relative pose between LiDAR-provided 3D point clouds and camera-provided images. These methods are primarily divided into two categories: learning-based and traditional matching-based. Learning-based algorithms, such as \cite{feng20192d3d}, place a greater focus on the front end, typically attempt to learn a 2D-3D matching relationship end-to-end. Such methods need to generate descriptors based on training sets or prior point cloud maps, which may lead to matching errors in unknown environments. 
Additionally, the errors in 2D-3D matching cannot be corrected in real-time, significantly impacting subsequent pose estimation. Traditional matching-based algorithms, such as \cite{gawel2016structure}, typically adopt structural descriptors to integrate sparse visual data with dense laser map. However, traditional methods usually rely on parameters and have difficulty solving generalization problems.
Both types of methods do not sufficiently emphasize the initialization of relative pose. Most algorithms face issues with relative pose initialization, either requiring significant computational time \cite{bai2023colmap}\cite{yu2020monocular}\cite{terzakis2020consistently} or necessitating manual provision of precise initial values \cite{ren2022corri2p}.

To address these challenges, we propose a coarse-to-fine framework to realize LiDAR-vision real-time collaborative localization. In this framework, by constructing the point-to-plane association of visual features and LiDAR points, the PSO-Sampling algorithm provides a coarse relative positioning. Then the precise positioning is achieved through constructing a Point-aided Bundle Adjustment (Point-aided BA) optimization structure. Besides, during the initialization process, a point cloud-aided camera Initialization framework based on multi-level sampling is used to obtain an accurate initial pose.
We conduct experiments to validate the effectiveness of the proposed algorithm. Both datasets and our own real-world experiments demonstrated that the algorithm exhibits robustness even with inaccurate initial pose estimations and can address the drift problem in monocular VIO in complex environments. Additionally, we extend the algorithm to multi-UAV scenarios and demonstrated its feasibility.

This work offers the following contributions: 
\begin{enumerate}


    \item Contribution 1: We propose a novel coarse-to-fine framework for LiDAR-vision real-time collaborative positioning. Within this framework, we utilize VIO to propagate the initial current camera pose and construct feature point clouds. Through pose sampling based on initial pose and an optimization backend using the spatial association of feature point clouds, we can achieve accurate localization.
    

    \item Contribution 2: We propose an adaptive event-triggered PSO-sampling algorithm to achieve relative accurate point cloud matching within a limited time interval.

    \item Contribution 3: We associate the feature point cloud and LiDAR point cloud based on 3D geometric spatial information and construct the point-aided BA problem based on point-to-plane residual.
    
    \item Contribution 4: We propose a multi-level sampling method that combines SFM feature points and LiDAR map to initialize the algorithm, addressing the challenge of lacking initial values.
    \item Contribution 5: We conduct the comparative experiments and validate the real-time performance, robustness, and accuracy of our algorithm on open source datasets and self-built datasets.
\end{enumerate}

\section{Related Works}
\label{sec:related-works}

\subsection{Visual-Inertial and LiDAR-Inertial Odometry}


Current LiDAR-Inertial Odometry (LIO) algorithms can achieve real-time pose estimation and precise 3D point cloud map construction \cite{xu2021fast,xu2022fast,legoloam2018shan,liosam2020shan}. Whether in filtering-based LIO algorithms or optimization-based LIO algorithms, the point-to-plane residual is widely used. This residual is utilized in Kalman filter observation equations \cite{xu2021fast}\cite{xu2022fast}, or as residual terms in factor graph optimization \cite{legoloam2018shan,liosam2020shan,shan2018lego}. When a new scan frame's LiDAR point cloud is received, it is first transformed into the LiDAR world frame. Then, the nearest points are used to fit local planes, and the point-to-plane distance residual is computed. 

Typical VIO systems, such as VINS\cite{qin2018vins} and OKVIS\cite{leutenegger2015keyframe}, are based on factor graph optimization\cite{kaess2008isam}. These systems take the reprojection error of visual feature points and the IMU preintegration error as residuals in the factor graph optimization. By minimizing these residuals, the optimal pose estimation is obtained. While these monocular VIO systems can achieve accurate and fast real-time pose estimation, issues such as drift over time and scale estimation errors still persist. 

\subsection{Learning-Based 2D-3D Matching}

For learning-based methods, establishing correspondences for registration through learning common feature descriptors is inherently challenging due to the absence of appearance and geometric correlations across the two modalities.
In \cite{feng20192d3d}, the authors propose an end-to-end deep network architecture to jointly learn the descriptors for 2D and 3D keypoint from image and point cloud respectively. Then based on this matching relationship, relative pose estimation is achieved. 
In \cite{li2021deepi2p}, the authors convert the registration problem into a classification and inverse camera projection optimization problem to avoid the problem of cross-modal association. In \cite{cattaneo2019cmrnet}\cite{chang2021hypermap}, the authors project map points onto a virtual image plane, and a synthesized depth image is generated using the intrinsic parameters of the camera and an initial pose estimate.

The above methods \cite{feng20192d3d}\cite{li2021deepi2p}\cite{wang2021p2,ren2022corri2p,jeon2022efghnet,cattaneo2020cmrnet++,chen2022i2d} always decompose the Image-to-PointCloud problem into two modules, 2D-3D associated module and pose estimation module.
However, errors in the 2D-3D associated module will not be corrected in subsequent pose estimation module because they are not jointly optimized for a unique goal. In \cite{wang2023end}, the authors achieve an fully end-to-end registration architecture (I2PNet) without separate modules and directly associates the 2D RGB image and the raw 3D LiDAR point cloud based on 2D-3D cost volume module. However, the The performance of I2PNet relies heavily high-quality LiDAR data. In environments where LiDAR data is sparse or noisy, the accuracy of the registration and localization may degrade.

\subsection{Traditional Matching-Based Relative Pose Estimation}

For traditional matching-based relative pose estimation algorithms, many works focus on finding a specific rule to establish matching relationships between 2D images and 3D point clouds. This matching relationship can be directly constructed between images and point clouds \cite{bai2023colmap}\cite{huang2020gmmloc,ye2020monocular,huang2019metric,Wolcott2014Urban}, or indirectly between sparse camera feature point map and dense LiDAR point map \cite{gawel2016structure}\cite{gawel20173d}\cite{li2020collaborative}.

In methods directly constructing matching relationships between images and point clouds, the 3D point cloud will be projected onto 2D image after some pre-processing. Then, specific similarity metrics are used to compare the similarity between the 3D point cloud and 2D image features to achieve matching. In \cite{bai2023colmap}, LiDAR points within the quadrangular pyramid in front of camera are projected onto the current camera imaging plane using the predicted camera pose, then a two-layer BA method is used to estimate camera pose. This process is quite time-consuming and can only be applied in offline scenarios. Unlike directly establishing matches between 2D points and 3D points, \cite{yu2020monocular} first extracts 2D and 3D line features and then establishes matching relationships between the line features. However, this approach requires extracting 3D line features from prior LiDAR maps and the extraction of 2D line features necessitates GPU-level computational power, which limits its applicability in real-world scenarios. Similar to the aforementioned work, \cite{yu2020monocular} also requires a prior map modeled by the Gaussian Mixture Model (GMM). During the matching step, after visual tracking, the Gaussian components are projected to the image coordinates and candidates are first searched in 2D. Then, with constraints from the map, the correspondence between local measurements and map components is established. This method also necessitates a prior map and uses GMM for map processing, rendering it unsuitable for dynamic point cloud scenarios.

In methods indirectly constructing matching relationships between sparse camera feature point clouds and dense LiDAR point clouds, structural descriptors are most commonly used. In \cite{gawel2016structure}, the authors adopt structural descriptors to merge sparse vision and dense laser maps. However, the descriptors utilized in this approach necessitate distinct parameters for various environments, lacking adaptability. In \cite{gawel20173d}, the authors extensively evaluate various methods for globally registering 3D point-cloud data generated by UGVs equipped with LiDAR sensors to point-cloud maps generated by UAVs with RGB-D sensors. Nevertheless, this approach necessitates UAV to preconstruct a large-scale point-cloud map.





    
\begin{figure*}[ht]
    \center
    \includegraphics[width=1\linewidth]{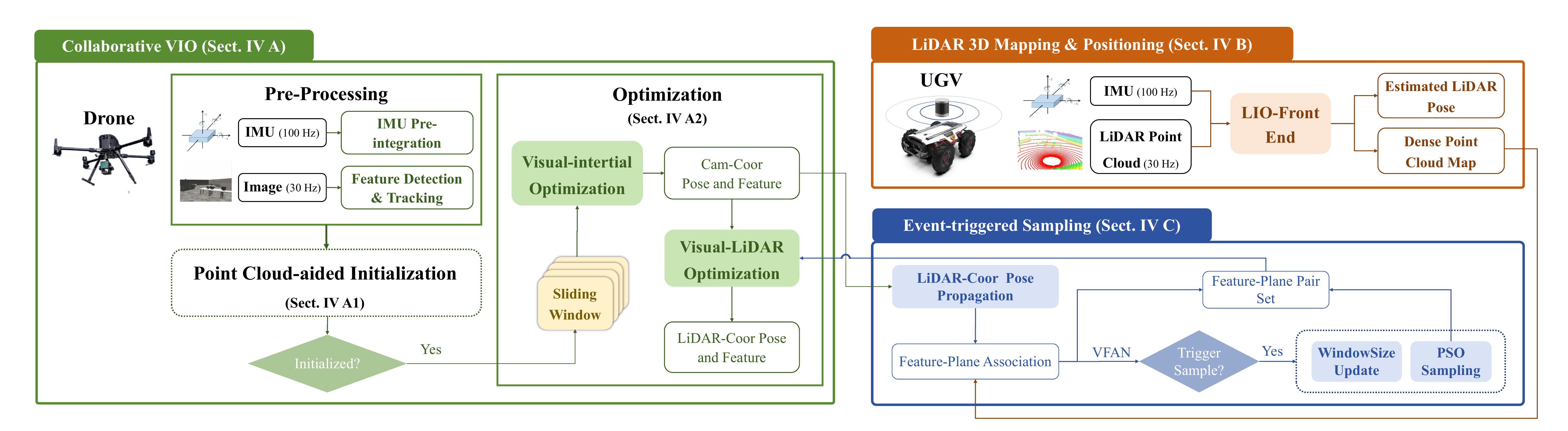}
    \caption{Workflow of tightly coupled air-ground Localization positioning system. The process integrates multiple data sources, including IMU, image, and LiDAR data, and is organized into several key modules: pre-processing, initialization, optimization, and event-triggered sampling.}
    \label{fig:sys-framework}
    \vspace{-0.3cm}
\end{figure*}

\section{Overview of the Framework}
\label{sec:framework}

\subsection{Problem Statement}
\label{subsec:problem}
\
Define $Im_{curr}$ as the current input image, $\mathcal{IM} =\left\{ Im_i\,\,|i=1\cdots N_I \right\} $ as the past $N_I$ frames of images, and $\mathcal{L} =\left\{ l_j\,\,| j=1\cdots N_L \right\} $ as the input LiDAR point set, where $l_j\in \mathbb{R} ^3$ represents the 3D coordinate of the LiDAR points. 

$Problem$: Given the current input image $Im_{curr}$, and the past image set $\mathcal{IM}$, the point map $\mathcal{L}$, find the the initial pose $T_{init}\in SE\left( 3 \right)$ of the first image, find the optimal current image pose $T_{curr}\in SE\left( 3 \right)$ and image poses set $\mathcal{T} =\left\{ T_i\in SE\left( 3 \right) \,\,| i=1\cdots N_I \right\} $ relative to the LiDAR coordinate system.

\subsection{Notation}

\begin{figure}[ht]
    \centering
    \includegraphics[width=1\linewidth]{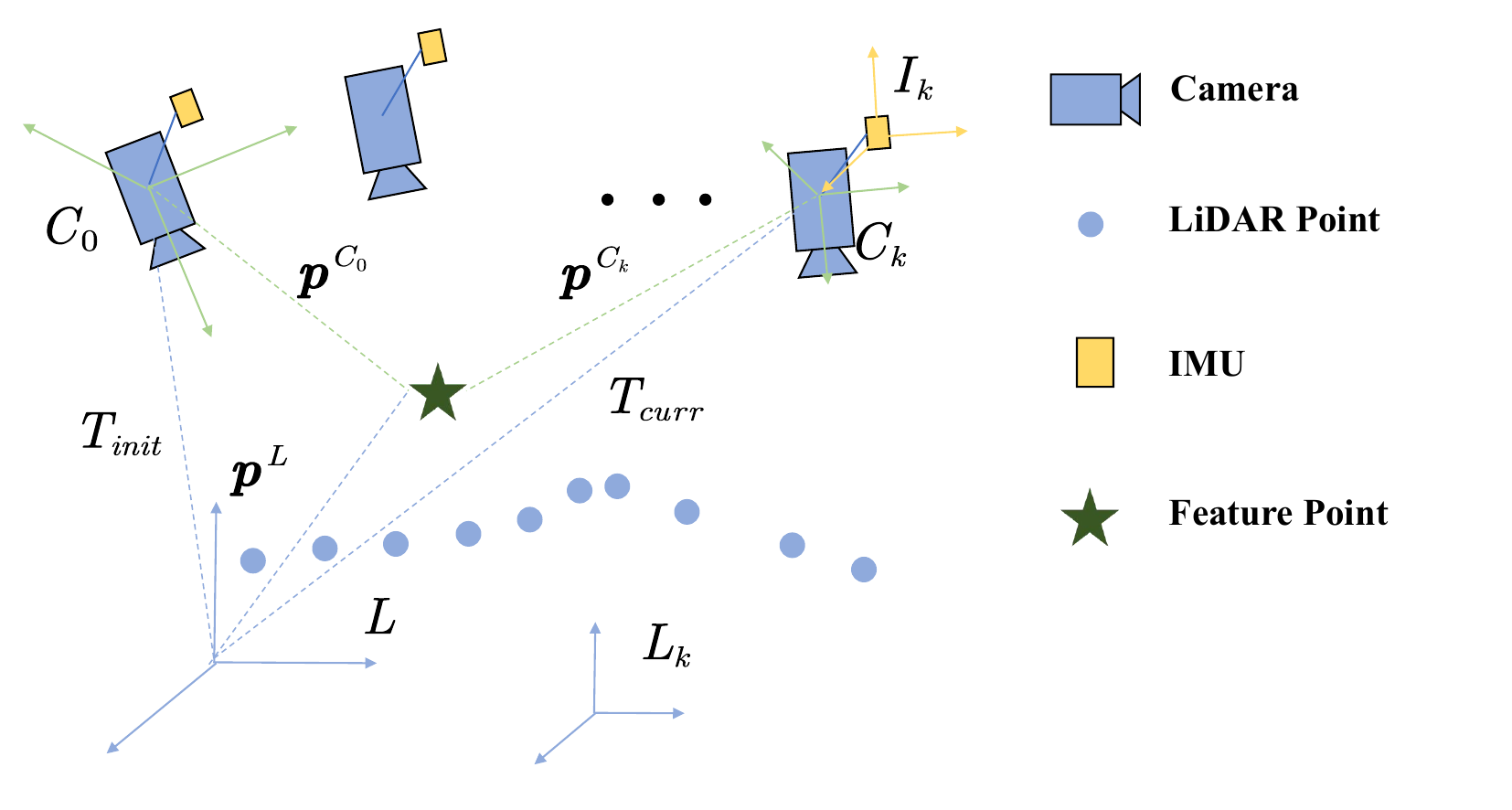}
    \caption{An illustration of the notations and coordinate transformations.
    }
    \label{fig:notation}
\end{figure}
We now specify notations and frame definitions used in this paper. We use $L$ for LiDAR-associated frmaes, $C$ for camera-associated frames and $I$ for IMU-associated frames. Frames related to LiDAR include LiDAR world frame and current LiDAR frame. We denote LiDAR world frame as $L$, which is constructed in the initialization step of LIO, usually considered to be in the first frame of the LiDAR. Current LiDAR frame at time $k$ is $L_k$. Frames related to camera contain initial camera world frame, current camera frame and IMU frame. Initial camera world frame is denoted as $C_0$ and current camera frame at time $k$ is $C_k$. IMU frame used in VIO at time $k$ is $I_k$. Point $\boldsymbol{p}$ in LiDAR world frame $L$ is $\boldsymbol{p}^L$ and in camera frame $C$ is $\boldsymbol{p}^C$. These frmaes are illustrated in Fig. \ref{fig:notation}.


\subsection{System Framework}
\ 
According to the sensor type and calculation process, our system is divided into three modules, namely Collaborative VIO module, LiDAR 3D Mapping \& Localization module, and Event-triggered Sampling module, as shown in Fig\ref{fig:sys-framework}. The Collaborative VIO module will be described in detail in Sec\ref{subsec:Collaborative VIO}, which is the main part of our system. In the module, Initialization, Optimization and Global Pose Graph Optimization part will be described respectively in SubSec\ref{subsec:Initialization}, SubSec\ref{subsec:Optimization} and SubSec\ref{subsec:Global Pose Graph Optimization}. The LiDAR 3D Mapping \& Localization module provides the Camera-LiDAR world coordinate system and relatively accurate LiDAR pose, and dynamically constructs geometric structure of the surrounding environment in the form of 3D point cloud map. This module will be described in Sec\ref{subsec:LiDAR 3d Mapping Positioning}. The Event-triggered Sampling module constructs the matching relationship between the sparse visual feature point cloud and the dense LiDAR point cloud in real time, which will be described in Sec\ref{subsec:Event-triggered Sampling}. We expand the positioning system from single drone to multiple drones in Sec\ref{subsec:Multi-agent Relative Positioning}.

.

\section{System Design}
\label{sec:implementation}

\subsection{Collaborative VIO}
\label{subsec:Collaborative VIO}

\begin{figure}[htp]
    \centering
    \includegraphics[width=1.05\linewidth]{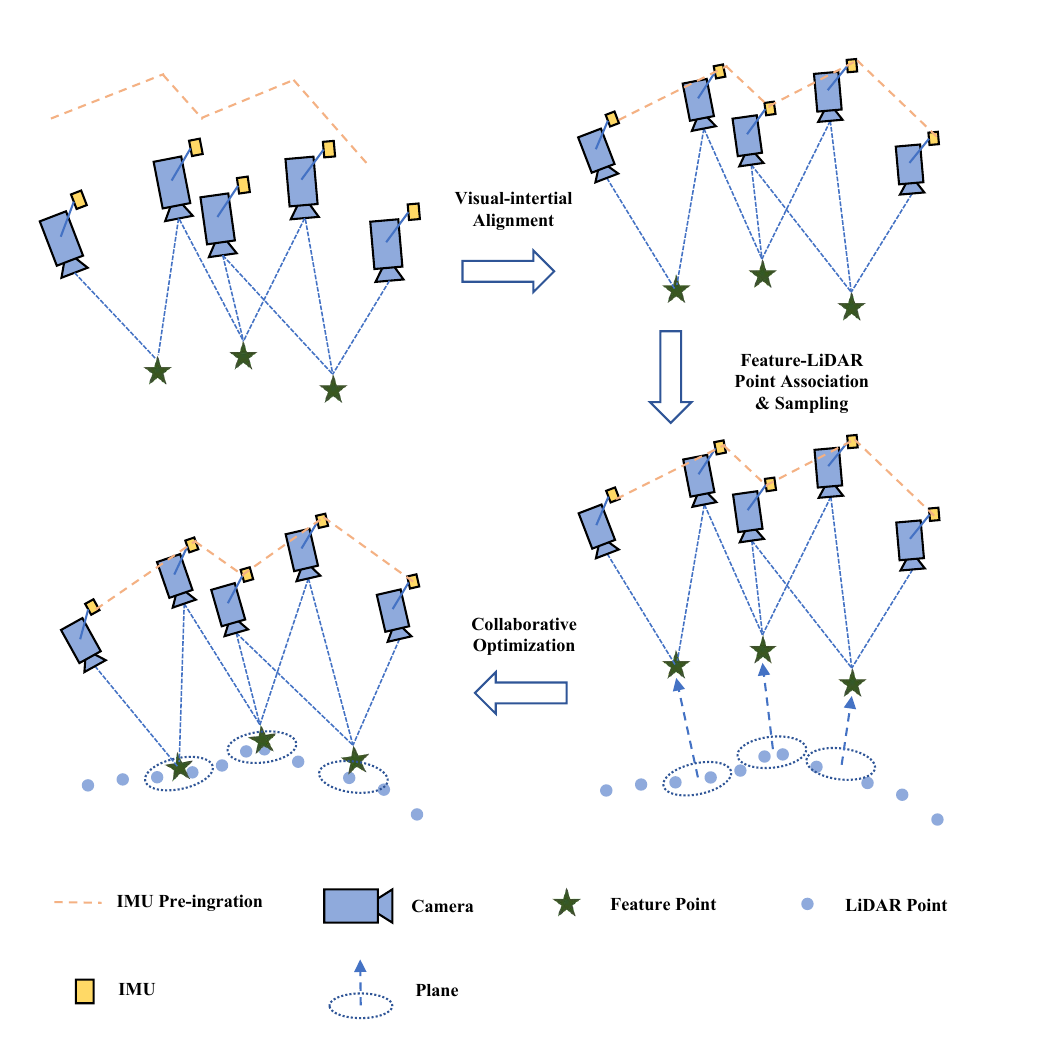}
    \caption{An illustration of the visual-inertial alignment and LiDAR-visual alignment processes for camera pose initialization.
    }
    \label{fig:frames}
\end{figure}

To correct VIO drift based on LiDAR point cloud, we propose the collaborative VIO. 
And this module is the main component of our LiDAR-Vision Collaborative Positioning (LVCP) system. This module processes real-time point clouds transmitted by the LiDAR 3D Mapping and Localization module and the coarse poses estimated by the Event-triggered Sampling module to construct a Point-aided BA problem to ensure precise localization in GPS-denied environments without prior maps. 

\subsubsection{Initialization}
\label{subsec:Initialization}
\
\newline
\indent In the initial process, we have no knowledge of the relative pose of the camera and LiDAR. The initial estimate of the relative poses needs to be conducted through the following measurements: the dense point cloud from the LiDAR and the sparse feature point cloud obtained by the camera. The basic idea of initialization is to estimate the relative pose through registering the 3D point clouds.

\begin{figure}[t]
    \centering
    \includegraphics[width=8.6cm]{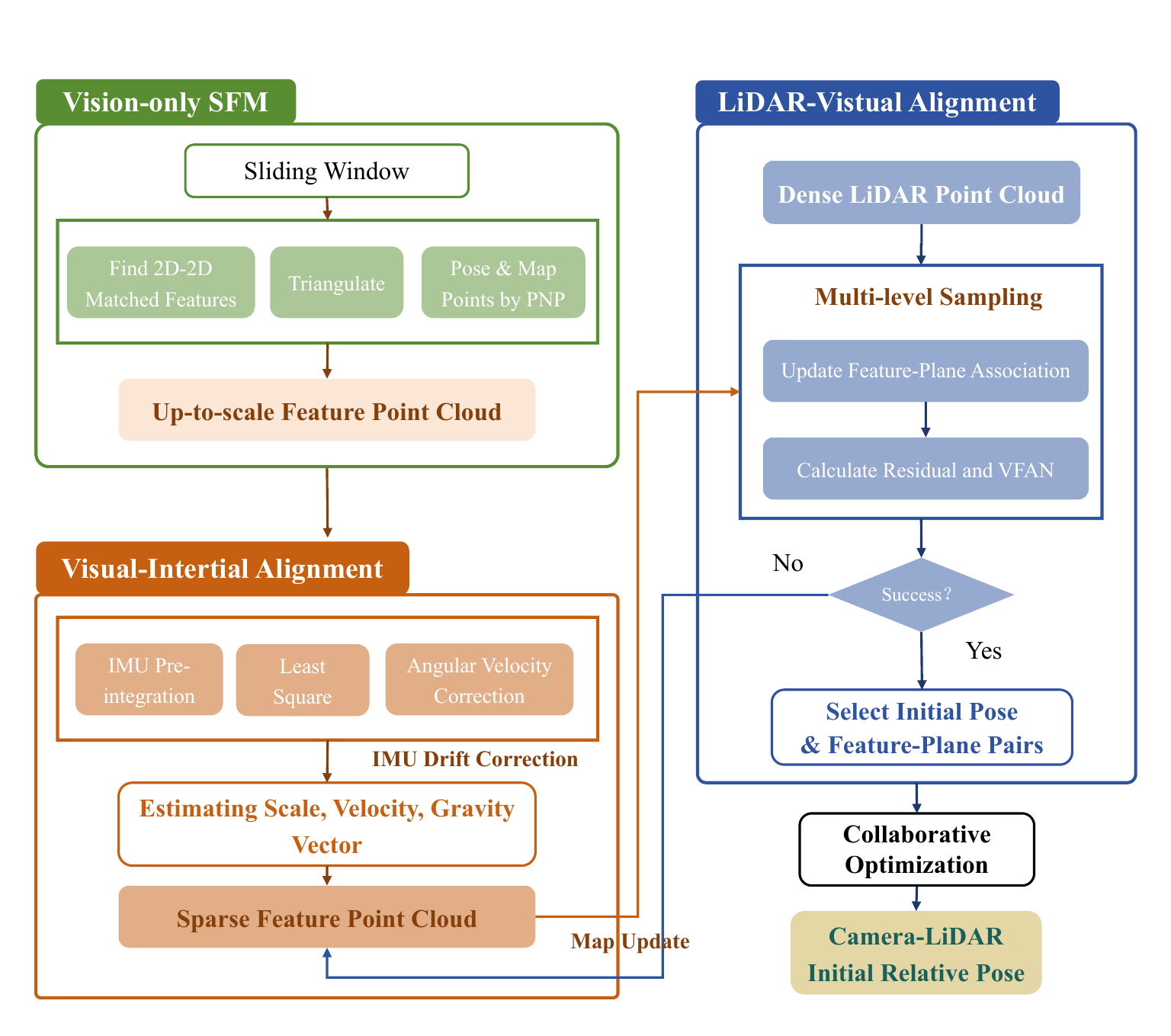}
    \caption{This figure illustrates the process of point cloud-aided camera initialization, divided into three moduals: Vision-only SFM, Vistual-Intertial Alignment and LiDAR-Vistual Alignment.}
    \label{Initialization}
    \vspace{-0.2cm}
\end{figure}


There exists mainly two challenges: 1) Due to the sparsity of point clouds generated by image information, spatial geometric features are difficult to characterize using descriptors. Therefore, the dense point cloud from the LiDAR cannot be directly matched point-to-point with the sparse feature point cloud; 2) Classic 3D point cloud registration algorithms, such as ICP\cite{121791}, require a precise initial value.

To solve the aforementioned two challenges, we propose a Point Cloud-aided Camera Initialization framework. In this architecture, sparse feature points are registered to LiDAR map based on Multi-level Sampling to estimate the rough initial camera pose, and then this pose is added to collaborative optimization to obtain an accurate initial pose. As shown in Fig.\ref{Initialization}, the visual front-end provides feature matching relationships between frames using KLT sparse optical flow algorithm \cite{lucas1981iterative}. We maintain a sliding window of frames with bounded computational complexity. The latest $10$ frames are used to build sliding window. 
After the operations of 2D-2D feature association, perspective-n-point (PnP) \cite{lepetit2009ep}, and triangulation method \cite{lepetit2009ep}, the sparse up-to-scale SFM point cloud is constructed. The visual-intertial alignment module uses the scale and orientation information generated by IMU motion to estimate the scale, velocity and gravity vector of the SFM point cloud \cite{qin2018vins}. The next step is based on sampling for coarse matching of feature point clouds and LiDAR point clouds, yielding initial pose and initial scale in the world (LiDAR) coordinate system.



\begin{figure*}[t]
    \center
    \includegraphics[width=18cm]{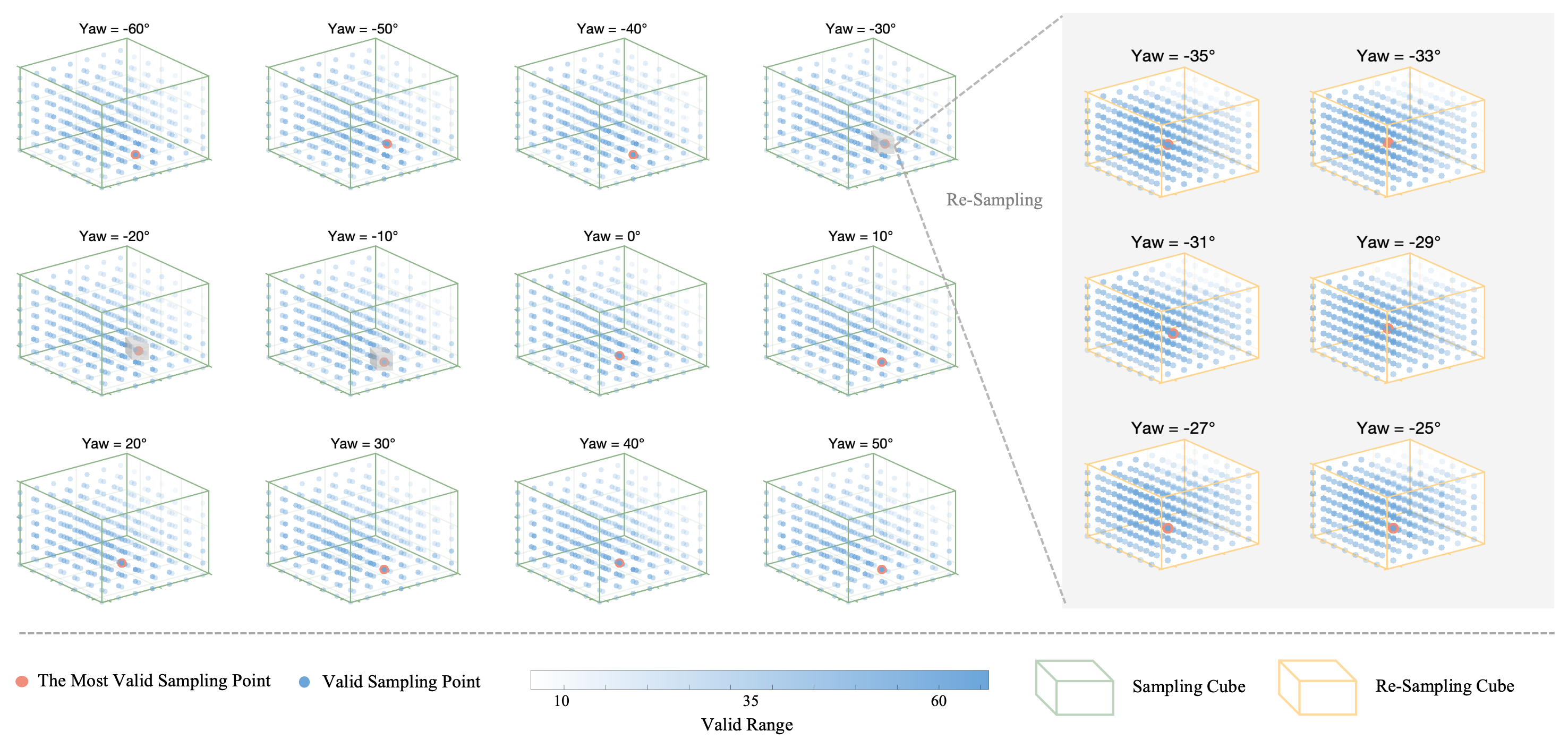}
    \caption{Multi-level Sampling.
    The green box on the left represents the 3D geometric space interval sampled in the first layer, and the different sampled boxes represent different rotations (in our experiment, since the roll pitch angle does not change much, we only sample the yaw), and the red box on the right represents part of the 3D geometric space interval sampled in the second layer. Each sampled subspace is sampled uniformly, and for each sampling point, the darker the blue, the more valid points there are. 
    }
    \label{fig:ampling}
    \vspace{-0.3cm}
\end{figure*}

The process of initialization sampling of camera pose is presented in Alg.\ref{alg:sample-init*}. The initial pose $\boldsymbol{x}_{init}:=[x_{init},y_{init},z_{init}]^T\in \mathbb{R} ^3$, yaw orientation $\theta _{init}$ and feature map scale $s_{init}$ are set to the sampling space $\mathcal{B}$. The following are the relevant definitions:
\begin{itemize}
    \setlength{\itemsep}{0pt}
    \setlength{\parsep}{0pt}
    \setlength{\parskip}{0pt}
    \item $\mathcal{F}$ represents camera feature point cloud and $\mathcal{F} :=\left\{ \boldsymbol{f}_m^{C_0}|m=1\cdots N_F \right\} $, where $\boldsymbol{f}_m^{C_0}\in \mathbb{R} ^3$ represents the 3D coordinate of the feature points in initial camera frame. 
    \item $ m _{Coarse}$ represents the number of valid feature points used to evaluate the sampling space.
    \item $\epsilon _{Coarse}$ represents the variance used to evaluate the sampling space.
    \item $ m _{Fine}$ represents the number of valid feature points used to evaluate the final initial sampling result.
    \item $\epsilon _{Fine}$ represents the variance used to evaluate the final initial sampling result.
\end{itemize}
$\mathcal{F}$ and $\mathcal{L}$ are set to be the input of Multi-level Sampling. And here are the definitions of the relevant functions:
\begin{itemize}
    \setlength{\itemsep}{0pt}
    \setlength{\parsep}{0pt}
    \setlength{\parskip}{0pt}
    \item $SampleSpaceBlocking(\mathcal{B})$:  Given the initial sampling space $\mathcal{B}$, divide the sampling space evenly into $S_c$ cube blocks, it returns the set $\left\{ \mathcal{B}_{i} \right\} _{i=1:S_c}$. 
    \item $FPAssociate\left( \left\{ \mathcal{B} _i \right\} _{i=1:S_c}, \mathcal{L}, \mathcal{F} \right)$:  Given the sampling sub-blocks $\left\{ \mathcal{B} _i \right\} _{i=1:S_c}$, feature point cloud $\mathcal{F}$ and LiDAR point cloud $\mathcal{L}$, for each $i$, the initial position $p$, pose $\theta$ and scale factor $s$ represented by the sampling space construct a corresponding plane $\varPhi _i$ based on $\mathcal{L}$ and $\mathcal{F}$. 
    After removing the outliers, each point $m$ of the feature point cloud in initial camera frame $\boldsymbol{f}_m^{C_0}$ is transferred to the LiDAR world frame $\boldsymbol{f}_m^L$:
    \begin{equation}
        \label{equ:f_m^W}
    	\begin{aligned}
    	\boldsymbol{f}_{m}^{L}=R\left( \theta \right) \boldsymbol{f}_{m}^{C_0}+\boldsymbol{x}_{init}
     \end{aligned}
    \end{equation}
    Based on KD-tree nearest neighbor search, find $5$ LiDAR points in $\mathcal{L}$ close to $\boldsymbol{f}_{m}^{L}$, and fit the local plane $\varPhi _{i,m}$ with normal vector $\boldsymbol{n}_m^L$ and its modular inverse $w_m$ using Householder QR decomposition \cite{trefethen2022numerical}. The detailed derivation is described in the Appendix.\ref{subsec:plane}. The plane $\varPhi _{i,m}$ corresponding to the $m$-th feature point under the $i$-th sampling space can be obtained.
    \item $FPEvaluate\left(\left\{\varPhi _{i,m} \right\} _{i=1:S_c,m=1:N_F} ,\mathcal{F}\right)$: Given the plane set $\{\varPhi _{i,m}\}_{i=1:S_c,m=1:N_F}$ and feature point cloud $\mathcal{F}$, it returns variance $\epsilon _i$ and valid matching number $m_i$. For each feature point, the point-to-plane distance can be calculated by Equ.\ref{equ:residual}. If this distance is less than the threshold, it will be considered a valid matching point. To mitigate the impact of outliers and noise, we calculate the variance within a $3\times 3\times 3$ neighborhood centered at $\mathcal B_{i}$. The presence of outliers typically results in an increased variance. $m_i$ and $\epsilon _i$ are used as indicators to evaluate the sampling effect in following steps.

    \item $Check\&Update\left( \left\{ \mathcal{B} _{i,j},
            m_{i,j},\tau _{i,j} \right\} \right)$:  Given the sub-sample space $\mathcal{B} _{i,j} $ and its corresponding $ m_{i,j}$ and $\epsilon _{i,j}$, combine the existing valid feature point number $m_{Best}$ and variance $\epsilon _{Best}$ to update the best sampling. 
   \end{itemize}

\SetKwFor{For}{for}{\string do}{}
\RestyleAlgo{ruled}
\begin{algorithm}[h]
    \caption{Multi-level Sampling($\mathcal{F}$,$\mathcal{L}$)}
    \label{alg:sample-init*}
    \LinesNumbered
    $\left\{ \mathcal{B_{i}} \right\} _{i=1:S_c}\gets SampleSpaceBlocking(\mathcal{B})$\;
    $\left\{\varPhi _{i,m} \right\} _{i=1:S_c,m=1:N_F}\gets FPAssociate\left( \left\{ \mathcal{B} _i \right\} _{i=1:S_c}, \mathcal{L}, \mathcal{F} \right) $\;
    \For{\rm \textit{i=} $1$ to $S_c$}
    {
        $\left\{ m_i,\epsilon  _i \right\}\gets FPEvaluate\left(\left\{ \mathcal{B} _i,\varPhi _{i,m} \right\}_{m=1:N_F} ,\mathcal{F}\right)  $\;
        \uIf{$\epsilon  _i<\epsilon  _{Coarse}\,\,\&\,\,m_i>m_{Coarse}$}{
        $\left\{ \mathcal{B} _{i,j} \right\} _{i=1:S_i}\gets SampleSpaceBlocking(\mathcal{B} _i)$\;
        $\left\{\varPhi _{i,j,m} \right\} _{j=1:S_i,m=1:N_F}\gets FPAssociate\left( \left\{ \mathcal{B} _{i,j}\right\} _{i=1:S_i}, \mathcal{L}, \mathcal{F} \right) $\;
        \For{\rm \textit{j=} $1$ to $S_i$}
        {
            $\left\{ m_{i,j},\epsilon  _{i,j} \right\}=FPEvaluate\left( \left\{ \mathcal{B} _{i,j},\varPhi _{i,j,m} \right\} _{m=1:N_F} ,\mathcal{F}\right) $\;
            $\left\{ \mathcal{B} _{best},
            m_{best},
            \epsilon  _{best} \right\} \gets Check\&Update\left( \left\{ \mathcal{B} _{i,j},
            m_{i,j},\tau _{i,j} \right\} \right) $\;
        }
        }
            
    }
    \uIf{$\epsilon  _{best}<\epsilon  _{Fine}\,\,\&\,\,m_{best}>m_{Fine}$}{
    \KwRet $\mathcal{B} _{best}$;
    }
    \uElse
    {
    \KwRet $Failure$;
    }
\end{algorithm}



As show in Fig.\ref{fig:ampling}, in our experimental scenario, the initial space is first divided evenly. 
The 3D geometry sampling interval of the first level is set to be $4\times 4\times 4m$, the resolution is $0.5m$, the yaw sampling interval is $120$ degrees, the resolution is $10$ degrees, and three sampling subspaces are screened out as the center points of the second level sampling. 
The 3D geometry sampling interval of the second level is set to be $0.5\times 0.5\times 0.5m$, the resolution is $0.0625m$, the yaw sampling interval is $12$ degrees, and the resolution is $2$ degrees. After the first-level sampling process, the second-level sampling space can obtain more valid points, which helps to find the most suitable initial camera pose faster. Experimental verification shows that under the same initial space and resolution condition, the double-level sampling time is only $0.1\%$ of the uniform sampling time, which greatly accelerates the initialization. 

\subsubsection{Collaborative Optimization}
\label{subsec:Optimization}
\
\newline
\indent 
Based on the initialization of multi-level sampling and real-time PSO-based sampling, which will be introduced in Sec.\ref{subsubsec: Real-time Sampling}, we can get a coarse camera pose in world coordinate system. In order to get more accurate value, we use BA optimization framework\cite{triggs2000bundle} to formulate our LiDAR-visual-inertial relative pose estimation problem. The premise of introducing LiDAR point cloud into the BA problem is that the pose of feature point cloud frame has a good initial estimation. Therefore, we divide backend optimization into two steps: Visual-Intertial optimization and Visual-LiDAR-Intertial optimization. 

In BA or factor graph optimization framework, we wish to maximize the likelihood of the measurements
$\mathcal{Z}$ given the history of system states and environment landmarks $\mathcal{X}$:
\begin{equation}
    \label{equ:max-likelihood}
    \mathcal{X}^\star=\arg\max_{\mathcal{X}}p(\mathcal{X}|\mathcal{Z})\propto p(\mathcal{X}_0)p(\mathcal{Z}|\mathcal{X})
\end{equation}
In Visual-Intertial optimization, measurements $\mathcal{Z}$ contains IMU
pre-integration measurements and visual feature points. In  Visual-LiDAR-Intertial optimization, beyond IMU and visual data, measurements $\mathcal{Z}$ also includes association between visual feature points and local plane in LiDAR point cloud. The aforementioned measurements are assumed to be conditionally independent and corrupted by white Gaussian noise, Eq.\ref{equ:max-likelihood} can be formulated as a weighted least squares minimization problem.

\textbf{\textit{Visual-intertial Optimization}}
\
\newline
\indent For Visual-Intertial optimization part, the full system state vector in optimization problem is defined as:
\begin{equation}
\label{equ:VIO}
\begin{aligned}
    & \mathcal{X}=\{\boldsymbol{x}_{c},\boldsymbol{x}_{f}\} \\ & 
    \boldsymbol{x}_{c}=\{[\boldsymbol{R}_{C_0C_k},\boldsymbol{t}_{C_0C_k},\boldsymbol{v}_{C_0C_k},\boldsymbol{b}_{a},\boldsymbol{b}_g]_{k=[0,n_{h}]}\} \\ &
    \boldsymbol{x}_{f}=\{\lambda_0,...,\lambda_{n_f}\}
    \end{aligned}
\end{equation}
where $C_0$ is initial camera world frame, $\{C_k\}_{k=[0,n_{h}]}$ represent all camera frames within sliding window, $n_h$ and $n_f$ are the number of camera frames and feature points in sliding window. $\boldsymbol{x}_{c}$ represents all states related to camera, including the camera pose $[\boldsymbol{R}_{C_0C_k},\boldsymbol{t}_{C_0C_k}]$, camera speed $\boldsymbol{v}_{C_0C_k}$ and imu bias of each frame within sliding window. $\boldsymbol{x}_{f}$ contain all states about feature points, specifically, all feature points' inverse depth in sliding window. $\boldsymbol{R}_{C_0C_k}$, $\boldsymbol{t}_{C_0C_k}$ and $\boldsymbol{v}_{C_0C_k}$ are rotation, translation and velocity of $C_k$ frame relative to $C_0$ frame. $\boldsymbol{b}_{a}$ and $\boldsymbol{b}_g$ represent acceleration bias and gyroscope bias.
$\lambda_l$ is the inverse depth of the $l^{th}$ feature in its first observation.

In Visual-Intertial graph factor optimization framework, we minimize the sum of two residuals: camera reprojection residual and imu pre-integration residual. Each term in following is the squared residual error associated to a factor type, IMU pre-integration factor and visual reprojection factor, weighted by the inverse of its covariance matrix.
\begin{equation}
    \begin{aligned} 
    & \sum_{k=0}^{n_{\mathcal{I}}-1}||\boldsymbol{\mathcal{R}}_{\mathcal{I}_{k,k+1}}(\boldsymbol{x}_{c},\mathcal{Z}_{I_k}^{I_{k+1}})||_{\Sigma^{-1}_{\mathcal{I}_{k,k+1}}} \\ &+ \sum_{i=0}^{n_{\mathcal{F}}}\sum_{j\in \mathcal{N}_{i}}||\boldsymbol{\mathcal{R}}_{\mathcal{C}_{ij}}(\boldsymbol{x}_{c},\boldsymbol{x}_{f})||_{\Sigma^{-1}_{\mathcal{C}_{ij}}}  \ ]
    \end{aligned}
\end{equation}
where $\mathcal{I}_{k,k+1}$ represents the pre-integration process between $k^{th}$ and ${k+1}^{th}$ IMU frame, and $n_{\mathcal{I}}$ is the number of pre-integration processes in sliding window. $\mathcal{Z}_{I_k}^{I_{k+1}}=\{\hat{\boldsymbol{\alpha}}_{I_k}^{I_{k+1}},\hat{\boldsymbol{\beta}}_{I_k}^{I_{k+1}},\hat{\boldsymbol{\gamma}}_{I_k}^{I_{k+1}}\}$ represents the pre-integration measurement between $k^{th}$ and ${k+1}^{th}$ IMU frame, and $\Sigma_{\mathcal{I}_{k,k+1}}$ is its covariance matrix. $n_{\mathcal{F}}$ is the number of feature points and $\mathcal{N}_{i}$ is the set contains all frames that can observe feature point $i$.

\textbf{IMU Pre-integration Factor}

Between two consecutive IMU frames, we pre-integrate IMU measurements about acceleration and angular velocity to provide pose and velocity constrain between two consecutive nodes of the factor graph. Consider two consecutive IMU frames $I_k$ and $I_{k+1}$ in the window, term $\boldsymbol{\alpha} $, $\boldsymbol{\beta}$ and $\boldsymbol{\gamma}$ respectively represent the pre-integration with respect to position, velocity, and angular velocity\cite{lupton2011visual}. Under frame $I_k$, the pre-integration of IMU measurements at discrete times $i$ and $i+1$ can be expressed the following. the estimated values are marked by $\hat{\left[ \cdot \right] }$.
    \begin{equation}
    \label{equ:abc}
	\begin{aligned}
        & \hat{\boldsymbol{\alpha}}_{i+1}^{I_k}=\hat{\boldsymbol{\alpha}}_{i}^{I_k}+\hat{\boldsymbol{\beta}}_{i}^{I_k}\delta t+\frac{1}{2}R\left( \hat{\boldsymbol{\gamma}}_{i}^{I_k} \right) \left( \hat{\boldsymbol{a}}_i-\boldsymbol{b}_{a_i} \right) \delta t^2
        \\&
        \hat{\boldsymbol{\beta}}_{i+1}^{I_k}=\hat{\boldsymbol{\beta}}_{i}^{b_k}+R\left( \hat{\boldsymbol{\gamma}}_{i}^{I_k} \right) \left( \hat{\boldsymbol{a}}_i-\boldsymbol{b}_{a_i} \right) \delta t
        \\ & \hat{\boldsymbol{\gamma}}_{i+1}^{I_k}=\hat{\boldsymbol{\gamma}}_{i}^{I_k}\otimes \left[ \begin{array}{c}
            	1\\
            	\left( \hat{\boldsymbol{\omega}}_i-\boldsymbol{b}_{\omega _i} \right) \delta t\\
            \end{array} \right] 
        	\end{aligned}
        \end{equation}
    where $\delta t$ is the time interval between two IMU measurements in VIO, we repeat the above recursive formula between $k^{th}$ frame and $k+1^{th}$ frame to get $\hat{\boldsymbol{\alpha}}_{I_k}^{I_{k+1}}$, $\hat{\boldsymbol{\beta}}_{I_k}^{I_{k+1}}$, and $\hat{\boldsymbol{\gamma}}_{I_k}^{I_{k+1}}$. $\otimes $ represents the multiplication operation between  two quaternions, $\hat{\boldsymbol{a}}_i$ and $\boldsymbol{b}_{a_i}$ represents the accelerometer measurement and its acceleration bias at time $i$, $\hat{\boldsymbol{\omega}}_i$ and $\boldsymbol{b}_{\omega _i}$ represents the gyroscope measurement and its bias at time $i$.

The IMU pre-integration residual is as following:
 \begin{equation}
 \label{equ:IMU factor}
        \begin{aligned}
            \begin{array}{l}
	\boldsymbol{\mathcal{R}} _{\mathcal{I}_{k,k+1}}(\boldsymbol{x}_c,\mathcal{Z} _{I_k}^{I_{k+1}})=\begin{pmatrix}
	\delta \boldsymbol{\alpha }_{I_{k+1}}^{I_k}\\
	\delta \boldsymbol{\beta }_{I_{k+1}}^{I_k}\\
	\delta \boldsymbol{\theta }_{I_{k+1}}^{I_k}\\
	\delta \mathbf{b}_a\\
	\delta \mathbf{b}_g\end{pmatrix}=\\
	\begin{pmatrix}
	\boldsymbol{R}_{C_0I_k}^{-1}( 
	\boldsymbol{p}_{C_0I_{k+1}}-\boldsymbol{p}_{C_0I_k}
	+\frac{1}{2}\boldsymbol{g}\Delta t_{k}^{2}-\boldsymbol{v}_{C_0I_k}\Delta t_k) -\boldsymbol{\hat{\alpha}}_{I_{k+1}}^{I_k}\\
	\boldsymbol{R}_{C_0I_{k+1}}^{-1}(
	\boldsymbol{v}_{C_0I_{k+1}}+\boldsymbol{g}\Delta t_k
	-\boldsymbol{v}_{C_0I_k}) -\boldsymbol{\hat{\beta}}_{I_{k+1}}^{I_k}\\
	2\left[ \boldsymbol{q}_{C_0I_k}^{-1}\otimes \boldsymbol{q}_{C_0I_{k+1}}\otimes \left( \boldsymbol{\hat{\gamma}}_{I_{k+1}}^{I_k} \right) ^{-1} \right] _{xyz}\\
	\boldsymbol{b}_{aI_{k+1}}-\boldsymbol{b}_{aI_k}\\
	\boldsymbol{b}_{wI_{k+1}}-\boldsymbol{b}_{wI_k}\\
\end{pmatrix}\\
\end{array}
        \end{aligned}
    \end{equation}
where $\boldsymbol{p}_{C_0I_k}$, $\boldsymbol{v}_{C_0I_k}$ and $\boldsymbol{q}_{C_0I_k}$ are estimated pose and velocity between camera initial frame $C_0$ and current IMU frame $I_k$. Continuous-time linearized dynamics of error terms can be derived to determine the covariance of residual $\Sigma_{\mathcal{I}_{k,k+1}}$, details are shown in Appendix \ref{subsec:IMU}.
    
\textbf{Visual Reprojection Factor}

    For the same feature point observed in multiple images, we construct the residual term by reprojection: 
    \begin{equation}
      \begin{aligned} \mathcal{\boldsymbol{R}}_{\mathcal{P}_{ij}}(\boldsymbol{x}_{c},\boldsymbol{x}_{f})= u_j -\pi_c(\frac{\boldsymbol{p}_{i}^{c_j}}{||\boldsymbol{p}_{i}^{C_j}||})
\\ \boldsymbol{p}_{i}^{C_j} = \boldsymbol{T}_{CI}\boldsymbol{T}_{C_0I_j}^{-1}\boldsymbol{T}_{C_0I_i}\boldsymbol{T}_{IC}\oplus 
 \frac{1}{\lambda}\pi_c^{-1}(u_i)
   \end{aligned} 
    \end{equation}
    where the certain camera feature point first observed by frame $i$, with the pixel coordinate $u_i\in \mathbb{R} ^2$, and its inverse depth is $\lambda \in \mathbb{R}$. It also can be observed by frame $j$, which pixel coordinate is $u_j \in \mathbb{R} ^2$. $\pi _c(u):\mathbb{R} ^3\rightarrow \mathbb{R} ^2$ is the projection function, which converts normalized coordinates in the camera coordinate system into pixel coordinates through the camera model. $\boldsymbol{T}_{CI}$ describes the fixed transformation from IMU to camera. $\oplus$ describes $SE(3)$ transform operation on vector.

\textbf{\textit{Visual-LiDAR Optimization}}
\
\newline
\indent After executing Visual-intertial Optimization, we can recursively obtain the relative pose of the current frame and the previous frame, and after PSO-Samping correction (Subsec.\ref{subsec:Event-triggered Sampling}), thereby providing a more accurate initial value for Visual-LiDAR Optimization. 

\begin{figure*}[ht]
    \center
    \includegraphics[width=18cm]{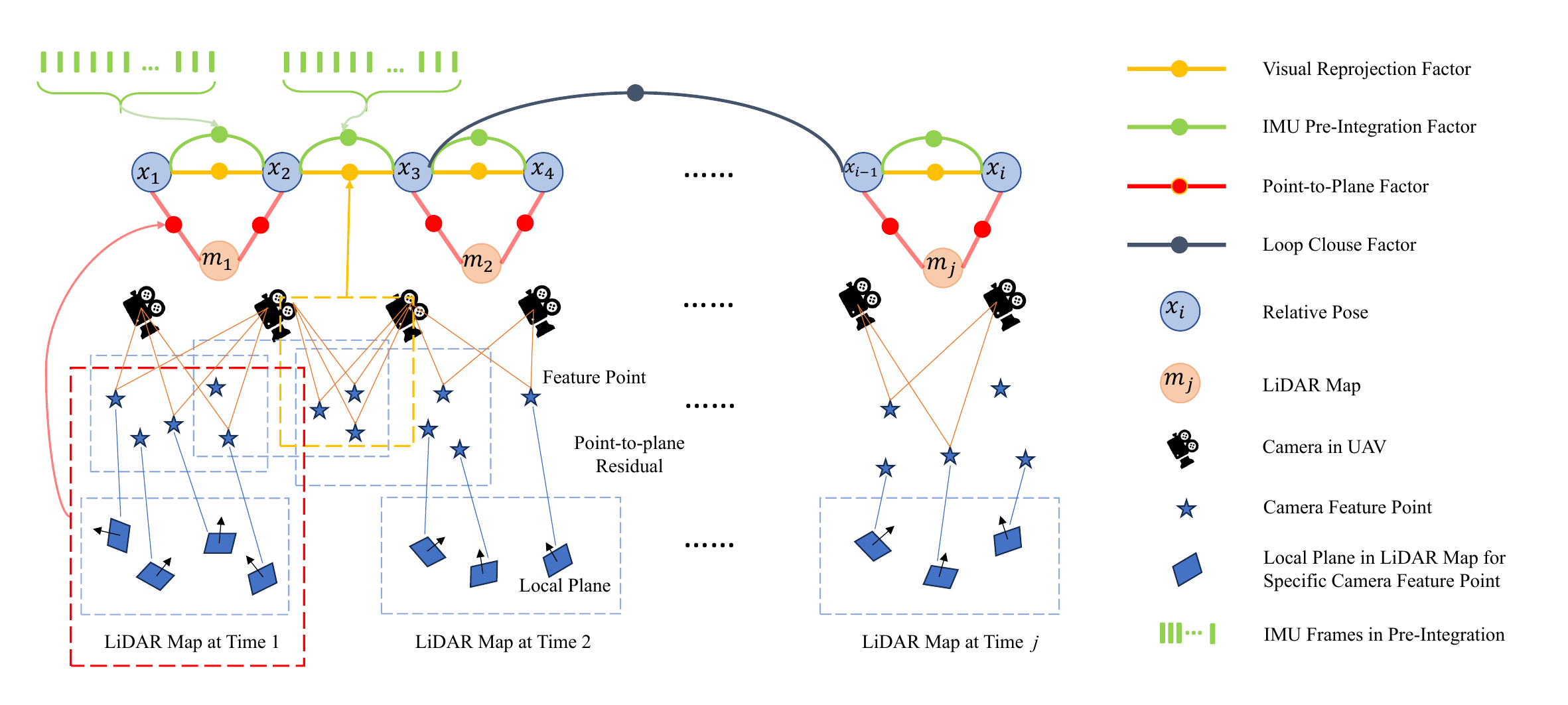}
    \caption{The system structure and factor graph of Our Method. Three types of data are received for system: LiDAR 3-D point cloud map, camera image and IMU measurement. Four Factors are constructed in optimization framework: 1) Point cloud point-to-plane factor, 2) Camera feature reprojection factor, 3) IMU pre-integration factor, and 4) Loop closure factor.
    }
    \label{fig:factor}
    \vspace{-0.3cm}
\end{figure*}


In single-agent SLAM, the system state estimation usually only focuses on the current camera pose. However, in LiDAR-Virtual Collaborative SLAM, the relative pose between the camera and LiDAR will be added as an optimization variable. This relative pose is defined as: $\boldsymbol{x}_{r}=[\boldsymbol{R}_{LC},\boldsymbol{t}_{LC}]$. Considering this relative pose, the system state and optimization variables are:
\begin{equation}
\begin{aligned}
    & \mathcal{X}_c=\{\boldsymbol{x}_{r},\boldsymbol{x}_{c},\boldsymbol{x}_{f}\} \\ & 
    \boldsymbol{x}_{r}=\{\boldsymbol{R}_{LC},\boldsymbol{t}_{LC}\} \\ &
    \boldsymbol{x}_{c}=\{[\boldsymbol{R}_{C_0C_k},\boldsymbol{t}_{C_0C_k},\boldsymbol{v}_{C_0C_k},\boldsymbol{b}_{a},\boldsymbol{b}_g]_{k=[0,n_{h}]}\} \\ &
    \boldsymbol{x}_{f}=\{\lambda_0,...,\lambda_{n_f}\}
    \end{aligned}
\end{equation}
where $L$ is LiDAR world frame, $C$ here refers to current camera frame, $C_0$ is initial camera frame, $\{C_k\}_{k=[0,n_{h}]}$ represent all camera frames within sliding window, $n_h$ and $n_f$ are the number of camera frames and feature points in sliding window. $\boldsymbol{x}_{r}$ are transformation from camera frame to LiDAR frame. $\boldsymbol{x}_{c}$ represents the camera pose, camera speed and imu bias of each frame within sliding window. $\boldsymbol{x}_{f}$ contains all feature points' inverse depth in sliding window. 

In LiDAR-Visual-Inertial relative pose optimization problem, we construct factor graph to minimize the weighted sum of three residuals: point cloud point-to-plane residual, camera reprojection residual, imu pre-integration residual.
\begin{equation}
    \begin{aligned}
    \mathcal{X}_c &=\mathrm{argmin}_{\mathcal{X}_c}[\ \sum_{l=0 }^{n_{\mathcal{P}}}||\boldsymbol{\mathcal{R}}_{\mathcal{P}_l}(\boldsymbol{x}_{r},\boldsymbol{x}_{f},\boldsymbol{p}_l)||_{\Sigma^{-1}_{\mathcal{P}_l}} \\ &+ \sum_{k=0}^{n_{\mathcal{I}}-1}||\boldsymbol{\mathcal{R}}_{\mathcal{I}_{k,k+1}}(\boldsymbol{x}_{c},\mathcal{Z}_{I_k}^{I_{k+1}})||_{\Sigma^{-1}_{\mathcal{I}_{k,k+1}}} \\ &+ \sum_{i=0}^{n_{\mathcal{F}}}\sum_{j\in \mathcal{N}_{i}}||\boldsymbol{\mathcal{R}}_{\mathcal{C}_{ij}}(\boldsymbol{x}_{c},\boldsymbol{x}_{f})||_{\Sigma^{-1}_{\mathcal{C}_{ij}}}  \ ]
    \end{aligned}
    \label{equ: point-to-plane residual}
\end{equation}
where $\mathcal{\boldsymbol{R}}_{\mathcal{P}_l}(\boldsymbol{x}_{r},\boldsymbol{x}_{f},\boldsymbol{n}_l)$ is the point-to-plane residual for point $l$ in point cloud matches between camera and LiDAR, $\mathcal{\boldsymbol{R}}_{\mathcal{I}_{k,k+1}}(\boldsymbol{x}_{c},\mathcal{Z}_{I_k}^{I_{k+1}})$ is the imu pre-integration residual between imu frame $k$ and $k+1$, $\mathcal{\boldsymbol{R}}_{\mathcal{C}_{ij}}(\boldsymbol{x}_{c},\boldsymbol{x}_{f})$ is the reprojection residual of same camera feature point between frame $i$ and $j$. $n_{\mathcal{P}}$ is the number of camera feature point and LiDAR point cloud local plane matchings, $n_{\mathcal{I}}$ is the number of imu frames within sliding window, $n_{\mathcal{F}}$ is the number of camera feature points in the sliding window. The factors are shown in Fig \ref{fig:factor}. Ceres Solver \cite{Agarwal_Ceres_Solver_2022} is used for solving this nonlinear optimization problem.

\textbf{Point Cloud Point-to-Plane Factor}

    We assume all feature points extracted by camera are correspond to geometric primitive in LiDAR world frame, specifically local planes. Surface features are usually composed of a large number of points and are less affected by single point noise. Therefore, when the amount of data is large and the noise is high, the results of surface feature extraction are usually more stable \cite{feng2014fast}\cite{besl1988segmentation}.
    Thus we only extract local plane in LiDAR point cloud, as in \cite{xu2021fast}.
    Hesse normal form of plane \cite{wisth2021unified} is used in this paper. Hessian normal form parametrizes an infinite plane $\boldsymbol{p}_{HN}$ as a unit normal $\hat{\mathbf{n}}\in \mathbb{R}^{3}$ and a scalar $d$ representing its distance from the origin:
    \begin{equation}
        \boldsymbol{p}_{HN}=\left\{\langle\hat{\mathbf{n}},d\rangle\in\mathbb{R}^4\mid\hat{\mathbf{n}}\cdot(x,y,z)+d=0\right\}
    \end{equation}
    Given a point $\boldsymbol{x}_0$, its distance to plane $\boldsymbol{p}_{HN}$ in hessian normal form is:
    \begin{equation}
        D = \hat{\mathbf{n}}\cdot \boldsymbol{x}_0 + d
    \end{equation}
    To identify the local plane associated with a specific point, a KD-Tree is used to locate nearby points and these points are fitted to a local plane. We then calculate the normal vector parameters of this plane, details are provided in Appendix.\ref{subsec:plane}. This normal vector is utilized in constructing the residual terms and the Jacobian matrix. 

    For the current camera frame $C$, coordinates of feature points in normalized plane are initialized by triangulation with their inverse depth.
    Given the coordinate of a camera feature point in normalized plane of the current camera frame $C$, which we denote $\boldsymbol{x}^C$, with its inverse depth $\lambda$. It can be transfered into LiDAR world frame through the estimated relative pose: 
    \begin{equation}
        \boldsymbol{x}^L=\frac{1}{\lambda}\boldsymbol{R}_{LC}\boldsymbol{x}^C+\boldsymbol{t}_{LC}
    \end{equation}
    Its corresponding local plane is $\boldsymbol{p}_{L}=\left\{\langle\hat{\mathbf{n}}_L,d_L\rangle\in\mathbb{R}^4\right\}$. Then the point-to-plane residual from a specific camera feature point to its corresponding LiDAR local plane is:
    \begin{equation}
    \label{equ:point-plane-residual}
    \boldsymbol{\mathcal{R}}_{\mathcal{P}_l}(\boldsymbol{x}_{r},\boldsymbol{x}_{f},\boldsymbol{p}_L)=\hat{\boldsymbol{n}}_L\cdot(\frac{1}{\lambda}\boldsymbol{R}_{LC}\boldsymbol{x}^C+\boldsymbol{t}_{LC})+d_L
    \end{equation}
The Jacobian matrix of this residual is:
    \begin{equation}
    \label{equ:point-plane-jacobian}
    \begin{aligned}
        \mathcal{\boldsymbol{J}}_{\mathcal{P}_l} & =\begin{pmatrix}
 \frac{\partial \boldsymbol{\mathcal{R}}_{\mathcal{P}_l}}{\partial \boldsymbol{R}_{LC}}  & \frac{\partial \boldsymbol{\mathcal{R}}_{\mathcal{P}_l}}{\partial \boldsymbol{t}_{LC}}  &\frac{\partial \boldsymbol{\mathcal{R}}_{\mathcal{P}_l}}{\partial \lambda} 
\end{pmatrix} \\ & =\begin{pmatrix}
-\hat{\boldsymbol{n}}_L^T\boldsymbol{R}_{LC}[\boldsymbol{x}^C]_{\times}  & \hat{\boldsymbol{n}}_L^T  & -\frac{1}{\lambda^2}\hat{\boldsymbol{n}}_L^T\boldsymbol{R}_{LC}\boldsymbol{x}^C 
\end{pmatrix}
\end{aligned}
    \end{equation}
The operator $[\boldsymbol{v}]_{\times}$
converts a vector $\boldsymbol{v} = (v_1, v_2, v_3)^T \in \mathbb{R}^3$ into the skew-symmetric “cross-product matrix”

$$\lfloor\boldsymbol{v}\rfloor_\times=\begin{bmatrix}0&-v_3&v_2\\v_3&0&-v_1\\-v_2&v_1&0\end{bmatrix},$$

We observe that, in formation of Jacobian matrix of point-to-plane residual, $\partial \boldsymbol{\mathcal{R}}_{\mathcal{P}_l}/\partial \lambda$ contains $-1/\lambda^2$ part. As inverse of point depth, $\lambda$ typically exhibits small values. During the optimization solver iteration process, if the $\lambda$ approaches zero or even becomes negative, it can lead to instability of the solution. Therefore, we fix the value of $\lambda$ in the Jacobian matrix and treat feature points whose inverse depth changes significantly after optimization as outliers, which then be excluded. We set $\Sigma_{\mathcal{P}_l^{-1}}$ as 2-5 times of threshold of planar fitting error check (commonly 0.05-0.2m).

\subsubsection{Global Pose Graph Optimization}
\label{subsec:Global Pose Graph Optimization}
\
\newline
\indent In Global Pose Graph Optimization part, we first perform loop closure detection in the Visual-Inertial Optimization part. When a loop closure is detected, a loop closure factor is added to the optimization terms to correct the currently estimated camera pose, resulting in a new pose after the loop. Subsequently, in the Visual-LiDAR Optimization, we first transform the currently estimated $\boldsymbol{T}_{LC}$ through the relative changes in camera pose to the new $\boldsymbol{T}_{LC}$. Afterward, we perform sampling and BA optimization, which includes IMU Pre-integration Residual, Visual Reprojection Residual, and Point cloud point-to-plane residual as shown in Subsec.\ref{subsec:Optimization}.

\subsection{LiDAR 3D Mapping \& Localization}
\label{subsec:LiDAR 3d Mapping Positioning}

The LiDAR 3D Mapping \& Positioning system consists of two parts: LiDAR-Inertial Odometry and 3D point cloud mapping. The LiDAR-Inertial Odometry method uses LiDAR point cloud and IMU data to estimate real-time pose of LiDAR, while 3D point cloud mapping method builds accurate map of surronding environment. This ploint cloud map will provide a platform for cameras to localize themselves in this map.

In LiDAR-Inertial Odometry, we use filter-based method to ensure real-time performance and effectiveness of pose estimation, as many classic works \cite{xu2021fast} \cite{xu2022fast}. Though Kalman filter, the following system states will be estimated:
 $$\mathbf{x}_k =[\boldsymbol{R}_{LL_k}^T,\boldsymbol{t}_{LL_k}^T,\boldsymbol{v}_{LL_k}^T,\boldsymbol{b}_w^T,\boldsymbol{b}^T_a,\boldsymbol{g}^T]^T\in SO(3)\times \mathbb{R} ^{15}$$
where $\boldsymbol{R}_{LL_k}\in SO(3)$ and $\boldsymbol{t}_{LL_k}\in \mathbb{R} ^{3}$ are rotation and translation from LiDAR current frame to LiDAR initial world frame, $\boldsymbol{v}_{LL_k}\in \mathbb{R} ^{3}$ is velocity of LiDAR in world frame, $\boldsymbol{b}_w\in \mathbb{R} ^{3}$ and $\boldsymbol{b}_a\in \mathbb{R} ^{3}$ are IMU angular velocity and acceleration bias, $\boldsymbol{g}\in \mathbb{R} ^{3}$ represents the gravity vector.

Once receiving the latest scan of LiDAR point cloud, it will be firstly transformed from current LiDAR scan frame to LiDAR world frame. Then, using the previously constructed kd-tree and dense point cloud map, we calculate point-to-plane residuals. The construction process of local plane is same as in initialization sampling process. These residuals are used to correct the estimated values of the error state iterative Kalman Filter algorithm, resulting in an updated pose estimation and a dense point cloud map.

The map points are organized into a kd-Tree, which dynamically grows by merging a new scan of point cloud at the odometry rate. To keep the size of the point cloud within a reasonable range, we dynamically delete points that are far away from the current position \cite{xu2022fast}. With the estimated pose $T_k$ at time $k$, points in scan of $k$ will be projected into global LiDAR frame.

\subsection{Event-triggered Sampling}
\label{subsec:Event-triggered Sampling}
\subsubsection{LiDAR-Coor Camere Pose Propagation}
\label{subsubsec:LiDAR-Coor Camere Pose Propagation}
\
\newline
\indent Between two adjacent camera frames $C_{k-1}$ and $C_k$, a good initial value of relative pose $\boldsymbol{R}_{C_{k-1}C_k}$ $\boldsymbol{t}_{C_{k-1}C_k}$ need to be estimated for subsequent LiDAR-visual-inertial relative pose problems. In our algorithm, we adopt VIO to derive $\boldsymbol{R}_{C_{k-1}C_k}$. 
Denote the relative rotation and translation between the camera-inertial coordinate system of frame $k-1$, $C_{k-1}$, and the LiDAR coordinate system $L$ as $\boldsymbol{R}_{LC_{k-1}}$ and $\boldsymbol{t}_{LC_{k-1}}$, respectively. We can obtain the relative rotation and translation of the current frame $k$ camera coordinate system $C_k$ relative to the LiDAR coordinate system $L$, denoted as $\boldsymbol{R}_{LC_k}$ and $\boldsymbol{t}_{LC_k}$, through the following recursive relationship:
\begin{equation}
   \boldsymbol{R}_{LC_k} = \boldsymbol{R}_{LC_{k-1}} \cdot \boldsymbol{R}_{C_{k-1}C_k} 
\end{equation}
\begin{equation}
    \boldsymbol{t}_{LC_k} = \boldsymbol{t}_{LC_{k-1}} + \boldsymbol{R}_{LC_k} \cdot \boldsymbol{t}_{C_{k-1}C_k}
\end{equation}
Note that the camera-inertial coordinate system $C_k$ is not the classic camera coordinate system but a coordinate system with the x-axis pointing forward, the y-axis pointing left, and the z-axis pointing up, with the origin coinciding with the classic camera coordinate system.

In VIO, we have already estimated the relative translation and rotation between the camera-IMU frame $I$ at frame $k-1$ and frame $k$ relative to the initial camera coordinate system $C_0$: $\boldsymbol{R}_{C_0I_{k-1}}$, $\boldsymbol{t}_{C_0I_{k-1}}$ and $\boldsymbol{R}_{C_0I_k}$, $\boldsymbol{t}_{C_0I_k}$. We use these to derive the relative pose transformation from frame $k-1$ to frame $k$, denoted as $\boldsymbol{R}_{C_{k-1}C_k}$ and $\boldsymbol{t}_{C_{k-1}C_k}$:
\begin{equation}
    \boldsymbol{R}_{C_{k-1}C_k} = \boldsymbol{R}_{CI} \boldsymbol{R}_{C_0I_{k-1}}^{-1} \boldsymbol{R}_{C_0I_k} \boldsymbol{R}_{IC}
\end{equation}
\begin{equation}
    \boldsymbol{t}_{C_{k-1}C_k} = \boldsymbol{R}_{CI} \boldsymbol{R}_{C_0I_{k-1}}^{-1} (\boldsymbol{t}_{C_0I_k} - \boldsymbol{t}_{C_0I_{k-1}})
\end{equation}
where $\boldsymbol{R}_{CI}$ is the rotation from camera frame to IMU frame, which are Priori accurate extrinsic parameters.


\subsubsection{Real-time Sampling}
\label{subsubsec: Real-time Sampling} 
\
\newline
\indent In the process of Visual-LiDAR Optimization, the accuracy of point-to-plane association has a great impact on the point-to-plane residual.
To correct the drift caused by VIO propagation, we need a method to estimate the precise position, scale, and rotation based on the existing VIO propagation. 
Considering position, scale, and rotation as optimization spaces, and VFN/TFN as the corresponding mapping, the existence of local optima and outliers makes traditional optimization methods (gradient descent and Newton method) fail, and due to the real-time requirement, large-scale fine-grained sampling methods cannot be performed. To address the challenge, we propose the event-triggered Particle Swarm Optimization (PSO) based Sampling method. PSO is a swarm intelligence-based optimization algorithm that mimics the social behavior of bird flocks or fish schools to find the global optimum. Its primary characteristic is leveraging information sharing among individuals to accelerate the convergence process.

\SetKwFor{For}{for}{\string do}{}
\RestyleAlgo{ruled}

\begin{algorithm}[ht]
    \caption{PSO-based Sampling($\mathcal{F}$,$\mathcal{L}$,$T_{LC_k}$)}
    \label{alg:pso}
    \LinesNumbered
    $\left\{ \mathcal{B}_{i} \right\} _{i=1:S_c} \gets AdoptiveSampleSpace(\mathcal{B}, T_{LC_k})$\;
    $\left\{ \boldsymbol{p}_i, \boldsymbol{v}_i \right\} _{i=1:m} \gets InitializePSO(\left\{ \mathcal{B}_{i} \right\} _{i=1:S_c})$\;
    
    $\left\{ \boldsymbol{pbest}_i \right\} _{i=1:m} \gets \left\{ \boldsymbol{p}_i \right\}_{i=1:m}$\;
    $\boldsymbol{gbest} \gets \arg \min_{\boldsymbol{p}_i} \rho (\boldsymbol{p}_i)$\;

    \For{\textit{t=} $1$ to $G_{max}$}
    {
        \For{\textit{i=} $1$ to $m$}
        {

            $\left\{\varPhi _{i,m} \right\} _{m=1:N_F} \gets FPAssociate(\boldsymbol{p}_i, \mathcal{L}, \mathcal{F})$\;

            $\rho (\boldsymbol{p}_i) \gets FPEvaluate(\left\{ \boldsymbol{p}_i, \varPhi _{i,m} \right\} _{m=1:N_F}, \mathcal{F})$\;

            \If{$\rho (\boldsymbol{p}_i) < \rho(\boldsymbol{pbest}_i)$}
            {
                $\boldsymbol{pbest}_i \gets \boldsymbol{p}_i$\;
            }

            \If{$\rho (\boldsymbol{p}_i) < \rho (\boldsymbol{gbest})$}
            {
                $\boldsymbol{gbest} \gets \boldsymbol{p}_i$\;
            }
        }

        \For{\textit{i=} $1$ to $m$}
        {
            $r_1 \gets rand(0,1)$\;
            $r_2 \gets rand(0,1)$\;
            $\boldsymbol{v}_i \gets w \cdot \boldsymbol{v}_i + c_1 \cdot r_1 \cdot (\boldsymbol{pbest}_i - \boldsymbol{p}_i) + c_2 \cdot r_2 \cdot (\boldsymbol{gbest} - \boldsymbol{p}_i)$\;
            $\boldsymbol{p}_i \gets \boldsymbol{p}_i + \boldsymbol{v}_i$\;
        }
    }

    \Return $\boldsymbol{gbest}$\;
\end{algorithm}

\begin{figure}[htp]
    \centering
    \includegraphics[width=1.0\linewidth]{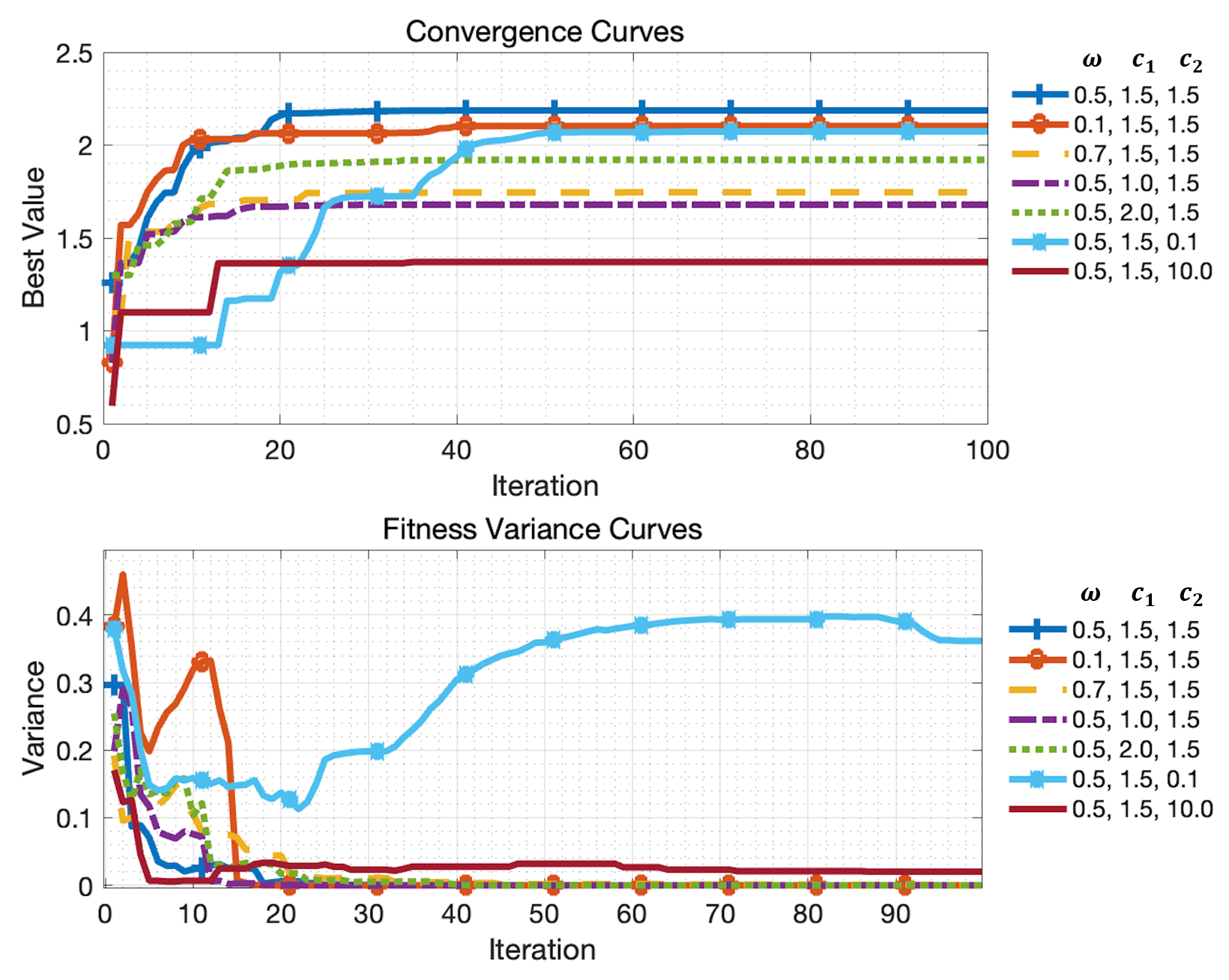}
    \caption{Fitness variance curves and convergence curves for different parameter combinations. Different lines stands for different parameter sets $\left\{ w,c_1,c_2 \right\} $ of PSO-based sampling method.}
    \label{fig:pso_para}
\end{figure}
The process of PSO-based sampling algorithm is presented in Alg.\ref{alg:pso}. The input pose is obtained by VIO propagation $\boldsymbol{T}_{LC_k}=\left[ \begin{matrix}
	\boldsymbol{R}_{LC_k}&		\boldsymbol{t}_{LC_k}\\
	0^T&		1\\
\end{matrix} \right] \in SE\left( 3 \right) $. We abbreviate valid feature point number as VFN and the total feature point number as TFN. The following are the relevant definitions:
\begin{itemize}
    \setlength{\itemsep}{0pt}
    \setlength{\parsep}{0pt}
    \setlength{\parskip}{0pt}
    \item $m$ represents the number of particles.
    \item $\left\{ \boldsymbol{p}_i,\boldsymbol{v}_i \right\} $ represents the $i-th$ particle's position and velocity.
    \item $\boldsymbol{pbest}_i$ represents the $i-th$ particle's personal best position.
    \item $\boldsymbol{gbest}_i$ represents the particle's global best position.
    \item $G_{max}$ represents iterations that the algorithm will execute.
    \item $\rho \left( \cdot \right) $ represents the fitness used to assess the quality of each particle's position.
    \item $w $ represents the inertia weight controls the influence of a particle's previous velocity on its current velocity. 
    \item $c_1 $ represents the personal learning coefficient determines the influence of the particle's personal best position $\boldsymbol{pbest}$ on its current velocity.
    \item $c_2 $ represents the social learning coefficient determines the influence of the global best position $\boldsymbol{gbest}$ on a particle's current velocity.
\end{itemize}
And here are the definitions of the relevant functions:
\begin{itemize}[]
    \setlength{\itemsep}{0pt}
    \setlength{\parsep}{0pt}
    \setlength{\parskip}{0pt}
    \item $AdoptiveSampleSpace(\mathcal{B},\boldsymbol{T}_{LC_k})$:  Given the initial sampling space $\mathcal{B}$, set $\boldsymbol{T}_{LC_k}$ as the center, divide the sampling space evenly into $S_c$ cube blocks. The smaller the ratio of VFN to TFN, the worse the correlation between the feature point cloud and the LiDAR point cloud, which will increase the size of the sampling space.
    \item $InitializePSO(\left\{ \mathcal{B}_{i} \right\} _{i=1:S_c})$:  Given the sampling space $\mathcal{B}_{i}$, randomly select $m$ points as the initial particles ${\boldsymbol{p}_i}$ and initialize their velocities ${\boldsymbol{v}_i}$.
    
\end{itemize}
We select a position in the running trajectory of the dataset V2-03-difficult to verify the influence of different parameter sets $\left\{ w,c_1,c_2 \right\} $ on the convergence of the PSO-based Sampling algorithm. In this comparative experiment, we set the number of particles $m$ to $10$ and the number of iterations $G_{max}$ to $100$. In Fig.\ref{fig:pso_para}, fitness variance is a measure of the dispersion of fitness values among all particles in the swarm during a particular iteration. A high fitness variance indicates that the particles are spread out over a wide range of fitness values, suggesting that the swarm is still exploring the search space.
When inertia weight $w$ increases, the algorithm becomes more difficult to converge. The increase of social learning coefficient $c_2$ will accelerate the group's convergence to the global optimal position, but this may lead to falling into the local optimal point.

\subsection{Multi-agent Relative Positioning}
\label{subsec:Multi-agent Relative Positioning}

\begin{figure}[t]
    \centering
    \includegraphics[width=1.0\linewidth]{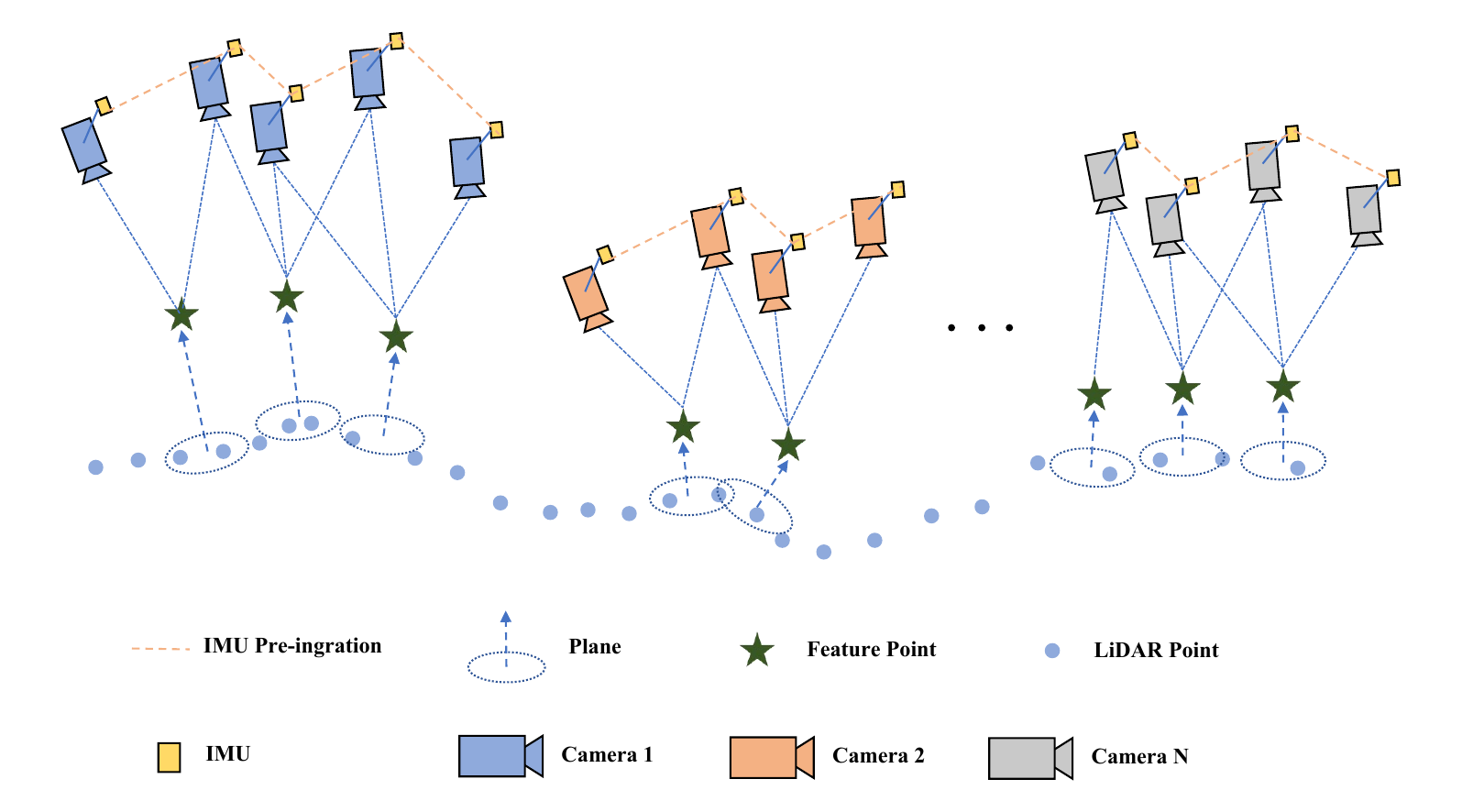}
    \caption{ This figure illustrates the extension of the collaborative BA optimization method to a multi-agent scenario. Multiple drones, each equipped with their own visual sensors, construct feature point clouds to localize their positions within a shared LiDAR point cloud provided by a UGV. 
    }
    \label{fig:multi}
\end{figure}

We extend the LVCP system to the multi-agent scenarios. As shown in the Fig.\ref{fig_exp_sen1} and Fig.\ref{fig:multi}, we introduce the following scenario, where a UGV equipped with a LiDAR builds a point cloud map in real time, and $N_{Drone}$ drones equipped with monocular cameras enter the point cloud map area one after another. Once entering the point cloud map area, the drone starts initialization process, and the LVCP system simultaneously builds the pose of all drones in the map area in the LiDAR coordinate system.


As shown in the Fig.\ref{fig:multi}, for each drone $i = 1:N_{Drone}$, the variables to be solved are defined as:

\begin{equation}
\begin{aligned}
    & \mathcal{X}_c^i=\{\boldsymbol{x}_{r}^i,\boldsymbol{x}_{c}^i,\boldsymbol{x}_{f}^i\} \\ & 
    \boldsymbol{x}_{r}^i=\{\boldsymbol{R}_{LC}^i,\boldsymbol{t}_{LC}^i\} \\ &
    \boldsymbol{x}_{c}^i=\{[\boldsymbol{R}_{C_0C_k}^i,\boldsymbol{t}_{C_0C_k}^i,\boldsymbol{v}_{C_0C_k}^i,\boldsymbol{b}_{a}^i,\boldsymbol{b}_g^i]_{k=[0,n_{h}]}\} \\ &
    \boldsymbol{x}_{f}^i=\{\lambda_0^i,...,\lambda_{n_f}^i\}
    \end{aligned}
\end{equation}
we construct the positioning problem as an independent Point-aided BA problem: 
\begin{equation}
    \begin{aligned}
    \mathcal{X}_c^i &=\mathrm{argmin}_{\mathcal{X}_c^i}[\ \sum_{l=0 }^{n_{\mathcal{P}}}||\mathcal{\boldsymbol{R}}_{\mathcal{P}_l}(\boldsymbol{x}_{r}^i,\boldsymbol{x}_{f}^i,\boldsymbol{n}_l)||_{\Sigma^{-1}_{\mathcal{P}_l}} \\ &+ \sum_{k=0}^{n_{\mathcal{I}}-1}||\mathcal{\boldsymbol{R}}_{\mathcal{I}_{k,k+1}}(\boldsymbol{x}_{c}^i,\mathcal{Z}_{I_k}^{I_{k+1}})||_{\Sigma^{-1}_{\mathcal{I}_{k,k+1}}} \\ &+ \sum_{i=0}^{n_{\mathcal{F}}}\sum_{j\in \mathcal{N}_{i}}||\mathcal{\boldsymbol{R}}_{\mathcal{C}_{ij}}(\boldsymbol{x}_{c}^i,\boldsymbol{x}_{f}^i)||_{\Sigma^{-1}_{\mathcal{C}_{ij}}}  \ ]\ \ ,\ \mathrm{for\  all}\ i
    \end{aligned}
\end{equation}
Note that for each drone, the LiDAR point cloud and the system's world coordinate system used to register features are shared. The accuracy of positioning depends on the accuracy of each drone's independent LVCP system and the quality of the shared LiDAR point cloud map (point cloud density, degree of point map structuring).

\section{EXPERIMENTS}
\label{sec:experiments}
\begin{table*}[ht]
\caption{\textbf{Qualitative Comparison of Tested Methods}}
\centering
\begin{tabular}{ccccccc}
\toprule
   & Line Corres & GMMLoc & Learning-Based & VINS-mono & Our Loose-Coupled & Our Tightly-Coupled
\\
\midrule 
need prior point cloud map ? &Y&Y&N&N&N&N \\
need GPU ? &Y&Y&Y&N&N&N \\
support dynamic point cloud map ?&N&N&Y&Y&Y&Y \\
can real-time running ? &Y&Y&Y&Y&Y&Y \\
need precise initial relative pose ?&Y&Y&Y&Y&N&N \\
\bottomrule
\end{tabular}
\label{tab:qualitative comparison}
\end{table*}

\begin{table*}[htp]
\caption{\centering \textbf{ATE RMSE Under Different Disturbance of Initial Translation (/m)}}
\centering
\begin{tabular}{cccccccccccccccc}
\toprule
  Dataset  &   \multicolumn{5}{c}{V1-03-difficult} & \multicolumn{5}{c}{V2-03-difficult} & \multicolumn{5}{c}{V2-02-medium}
   \\ \cmidrule(r){2-6} \cmidrule(r){7-11}  \cmidrule(r){12-16}
   Initial Disturbance (m) & 0 & 0.05 & 0.1 & 0.2 & 0.5 
     & 0 & 0.05 & 0.1 & 0.2 & 0.5
     & 0 & 0.05 & 0.1 & 0.2 & 0.5
\\
\midrule 
Line Corres & \textbf{0.21} & \textbf{0.25} & \textbf{0.27} & 0.46 & 0.77 & 0.89 & 0.91 & 0.81 & 1.00 & 1.06& 0.58 & 0.61 & 1.71 & 1.82 & 1.21\\
GMMLoc & 1.97 & 1.96 & 1.97 & 1.97 & 2.10  & failed & failed & failed & failed & failed & 2.00 & 2.00 & 2.00 & 2.00 & 6.85\\
VINS-mono (Loop) & 0.44 & 0.51 &  0.33 & 0.74 & 1.19 & 0.80 & 0.73 &  0.67 & 0.87 & 1.34 & \textbf{0.32} & \textbf{0.36} &  0.53 & 0.58 & 0.94\\
Our Loosely-Coupled & 0.41 & 0.37 &  0.45 & \textbf{0.38} & \textbf{0.45} & 0.41 & 0.39 &  0.35 & \textbf{0.37} & 0.47 & 0.38 & 0.43 &  0.41 & 0.42 & \textbf{0.35} \\
Our Tightly-Coupled & 0.35 & 0.43 & 0.40 & 0.49 & 0.51 & \textbf{0.39} & \textbf{0.36} & \textbf{0.29} & 0.41 & \textbf{0.41}
& 0.39 & 0.42 & \textbf{0.39} & \textbf{0.37} & 0.48\\
\bottomrule
\end{tabular}
\label{tab:ATE comparison1}
\end{table*}

We evaluate our algorithm on open source EuRoC MAV dataset \cite{burri2016euroc} and self-built datasets respectively. We first compare the LVCP system with other methods on the dataset to verify its accuracy and robustness in Subsec.\ref{subsec:Comparison}. Then, we build UGV and UAV platforms and build our own datasets to verify the effectiveness of our method in incremental point cloud map scenarios and multi-angent scenarios in Subsec.\ref{subssec:dynamic}. In order to verify the generalization, we conduct experiments in more scenarios in Subsec.\ref{subsec:Complex}.


\subsection{Comparison on EuRoC MAV Dataset}
\label{subsec:Comparison}
We first test our proposed method on EuRoC MAV dataset \cite{burri2016euroc}. EuRoC MAV dataset contains visual-inertial datasets collected on-board a Micro Aerial Vehicle. The MAV carrys a visual-inertial (camera-IMU) sensor unit, which can get Stereo Images (Aptina MT9V034 global shutter, WVGA monochrome, 2×20 FPS) and MEMS IMU data (ADIS16448, angular rate and acceleration, 200 Hz). Leica MS50 3D structure scan provided by dataset can serve as static 3D LiDAR pointcloud map. We use the real-time image on the left side of MT9V034 and the IMU as the sensor of the drone. The comparative experiments are conducted on a desktop computer with an Intel i7-12700 CPU and 16 GB of RAM. 
EuRoC MAV dataset provides pre-built static 3D LiDAR pointcloud map of the vicon room environment and groundtruth trajectories. We select sub-datasets "V1-03-difficult", "V2-03-difficult" and "V2-02-medium", where the running trajectories are more complex and the camera shaking is more obvious. The LVCP system does not require a priori point cloud map or pre-processing of the map. However, given that there is currently no reliable point-aided algorithm that can run in real time and does not require a priori knowledge like LVCP, we use a pre-built static map to conduct comparative experiments to verify the accuracy of the algorithm. In this case, 
the position of the LiDAR is considered to be at the origin of the world coordinate and will not move.


We compare our method with several algorithms on EuRoC MAV dataset, including VINS-mono\cite{qin2018vins}, 2D-3D line correspondences localization\cite{yu2020monocular}, and GMMLoc\cite{huang2020gmmloc}. To simplify the notation, we use Line Corres to denote 2D-3D line correspondences localization. 
Our method includes both loosely coupled and tightly coupled approaches. Loosely coupled means, in point cloud point-to-plane residual (\ref{equ:point-plane-residual}), we treat the inverse depth of each camera feature point as constant variable, i.e., the term $\partial \boldsymbol{\mathcal{R}}_{\mathcal{P}_l}/\partial \lambda$ in the Jacobian matrix (\ref{equ:point-plane-jacobian}) is set to zero. The tightly coupled approach uses LiDAR point cloud to correct inverse depth of camera feature points.

\begin{figure*}[htp]
    \centering
    \includegraphics[width=18cm]{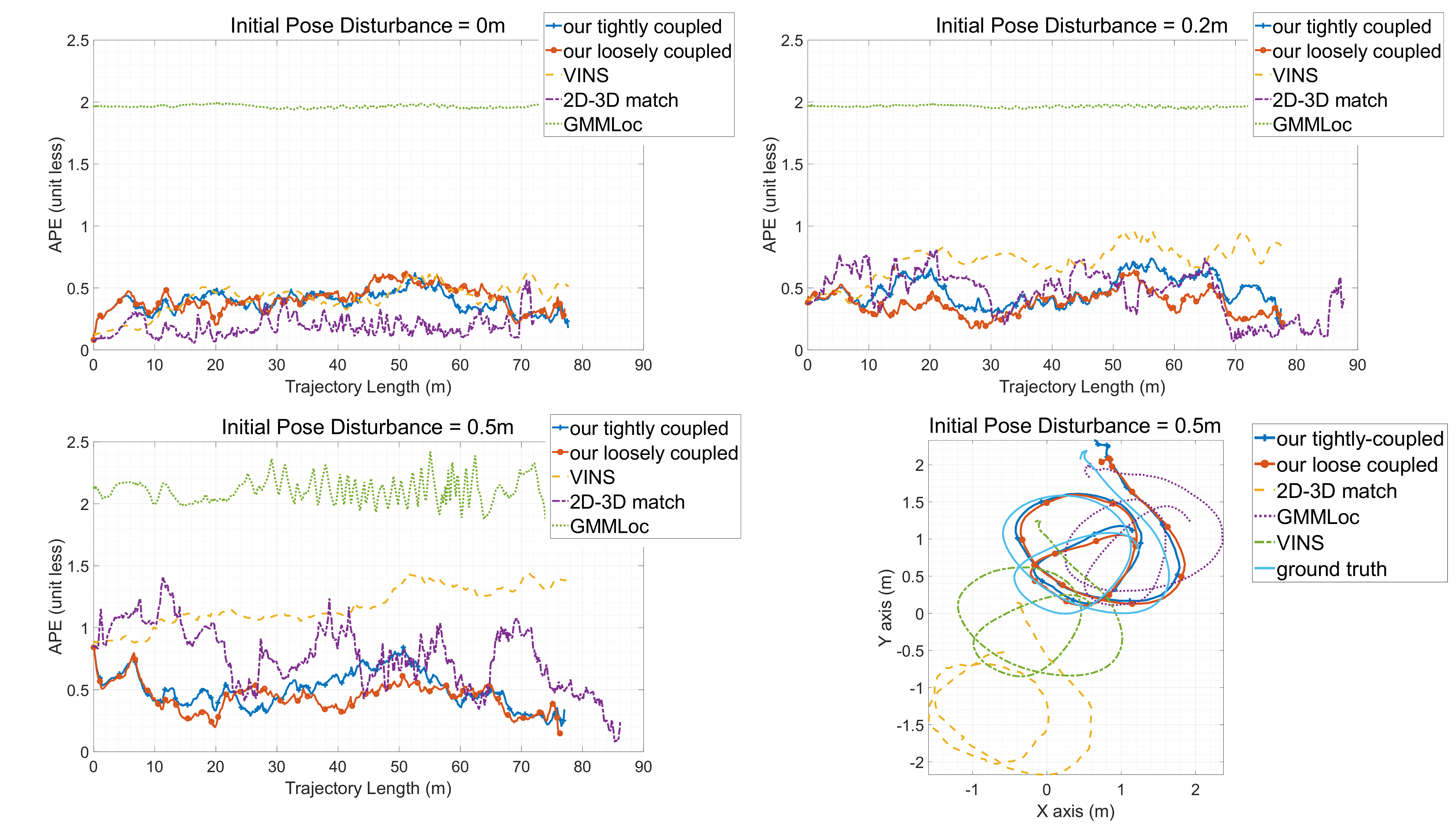}
    \caption{Trajectory errors under different initial pose disturbances in the "V1-03-difficult" dataset scenario. Figures (a), (b), and (c) show the RMSE of APE. It can be seen that as the initial pose disturbance increases, our method consistently maintains a low trajectory estimation error. Figure (d) displays the trajectory of the final segment, where it can be observed that our method continuously corrects the estimation drift.}
    \label{fig:trajs-v1-03}
\end{figure*}

Each method has been evaluated based on several key criteria that are critical for the effectiveness of UAV and UGV collaborative positioning systems. The criteria include the need for a prior point cloud map, the requirement for GPU, support for dynamic point cloud maps, real-time running capability, and the necessity of a precise initial relative pose. Our proposed methods, both loose-coupled and tightly-coupled, offer significant advantages by not requiring a prior point cloud map or precise initial relative pose while supporting dynamic point cloud maps and maintaining real-time performance. These features make our methods more adaptable and robust in varied and challenging environments, especially for applications involving multiple UAVs and UGVs.

It should be noted that in Line Corres method, geometric 3D lines are extracted offline from LiDAR maps, and in GMMLoc, the prior map needs to be modelled by the Gaussian Mixture Model. We perform the above operations before the online comparative experiment. Since Line Corres and GMMLoc need to obtain the initial camera pose in advance, in the comparative experiments, the initial values are all groundtruth and its perturbation value. We do not numerically compare our approach with learning-based methods since these methods are generally not open source and most cannot run in real time. The qualitative description of these methods is shown in TABLE \ref{tab:qualitative comparison}.  

We evaluate our algorithm and methods listed above on three sequences, "V1-03-difficult", "V2-03-difficult" and "V2-02-medium". All estimated poses of each sequence are used for trajectory alignment with ground-truth\cite{grupp2017evo}, and we calculate absolute trajectory error (ATE) \cite{sturm2012benchmark} for comparison. The ATE results of each methods are shown in TABLE \ref{tab:ATE comparison1} . When calculating ATE in TABLE \ref{tab:ATE comparison1}, we consider both translation and rotation part. 

Furthermore, we test the performance of different algorithms in the presence of disturbance. 
We apply gradually increasing perturbations to initial pose estimation of each algorithm, from precise initial pose (i.e. disturbance is 0m) to imprecise initial pose (i.e. disturbance is 0.5m). TABLE \ref{tab:ATE comparison1} shows ATE RMSE of different methods under initial pose estimation disturbance (here we do not perform automatic alignment of trajectories to reflect the absolute pose estimation error). 

As shown in TABLE \ref{tab:ATE comparison1}, Line Corres method shows increasing error with greater initial disturbances. GMMLoc fails under initial disturbances in some sequences. VINS-mono with loop closure shows moderate sensitivity to initial disturbances. Our proposed loosely-coupled method demonstrates stable performance across different disturbances. Our proposed tightly-coupled method shows stable performance and slightly better robustness compared to the loosely-coupled approach. This comparison illustrates the effectiveness of our methods in maintaining low ATE RMSE despite varying initial disturbances, highlighting their robustness and reliability in challenging environments. Fig.\ref{fig:trajs-v1-03} shows trajectory errors under different initial pose disturbances in the "V1-03-difficult" dataset scenario, which highlights the effectiveness of the proposed method in maintaining low trajectory estimation errors and correcting drift under various initial pose disturbances.

\begin{figure*}[htp]
    \centering
    \includegraphics[width=18cm]{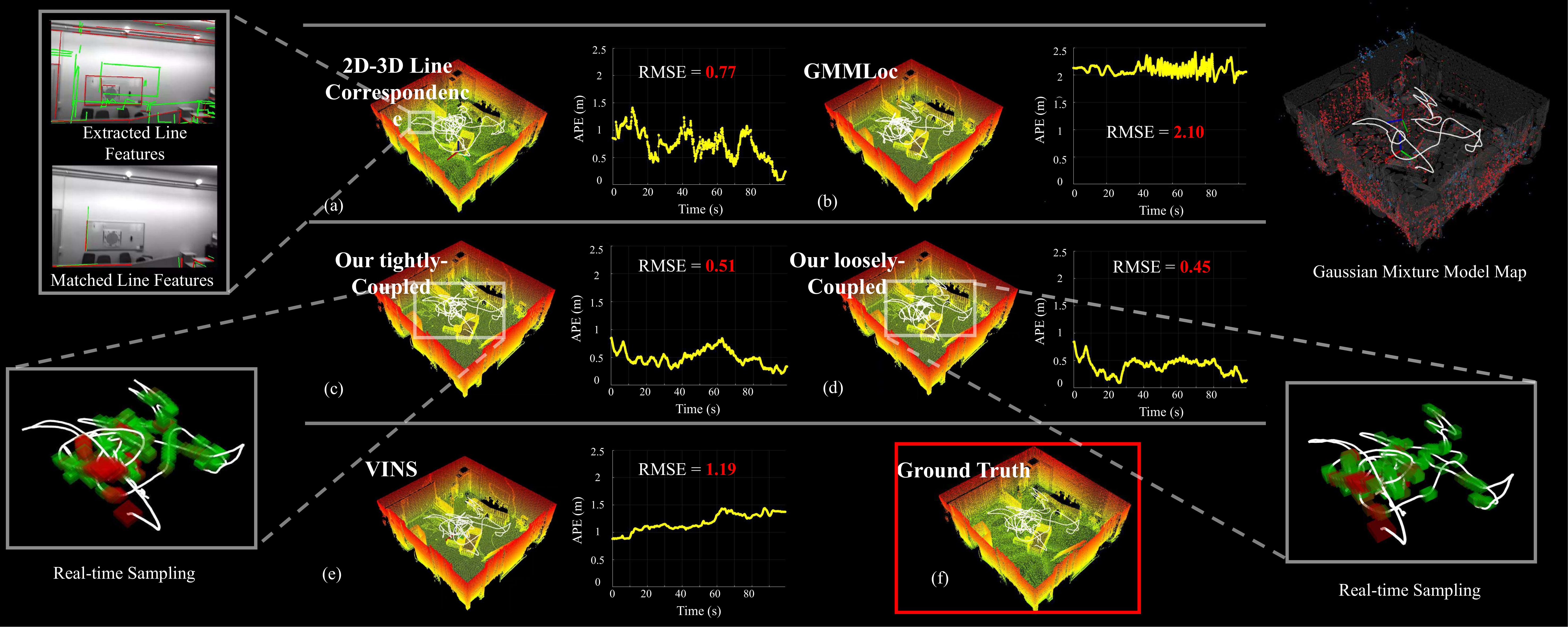}
    \caption{The estimated trajectories and errors in the "V1-03-difficult" scenario with an initial pose disturance of 0.5m. Figures (a)-(f) represent the estimated results of 2D-3D line correspondences, GMMLoc, our tightly coupled method, our loosely coupled method, VINS, and the ground truth.}
    \label{fig:all-data-v1-03}
\end{figure*}

\begin{figure}[htp]
    \centering
    \includegraphics[width=1.0\linewidth]{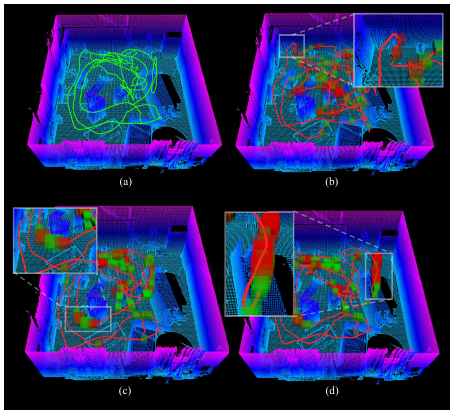}
    \caption{Figures (a), (b), (c) and (d) respectively show the results of running groundtruth, AUS, GRS, and PSO on the dataset V2-03-difficult. }
    \label{fig:psotest}
\end{figure}

\begin{table}[htp]
\centering
\caption{\textbf{Results of Tested PSO on EuRoC MAV Dataset}}
\label{table: PSO_test}
\tabcolsep=0.1cm
\begin{tabular}{lccccc}
\toprule
Method & Time (ms) & RMSE (m) & SR & ValidRatio & SN \\
\midrule 
AUS & 579 & 1.264 & \textbf{0.99} & \textbf{0.55} & 178 \\ 
GRS & \textbf{80} & 0.45 & 0.79 & 0.33 & 173 \\ 
PSO (Ours) & 119 & \textbf{0.383} & 0.79 & 0.32 & \textbf{169} \\ 
\bottomrule
\end{tabular}
\end{table}

In order to evaluate the efficiency of the real-time sampling algorithm PSO in correcting VIO drift, we designed a set of comparative experiments including greedy random sampling (GRS), and adaptive window uniform sampling (AUS). We adopt RMSE, single sampling execution time (Time), sampling success rate (SR), average valid point ratio (ValidRatio) as evaluation indicators and sampling number (SN). The SR is set to be the ratio of successful sampling number and total sampling number, and ValidRatio is set to be the average value of the ratio of valid feature number and total feature number for each frame. We use V2-03-difficult as the test dataset.

As shown in table.\ref{table: PSO_test}, The RMSE value of AUS is too high, indicating that it is far away from the grundtruth. This is because there may be outliers in the sampling space as shown in Fig.\ref{fig:psotest} (b). Outliers are points that do not cluster in a certain area of the sampling space like inliers, and thus do not have local convexity, and are considered to be an incorrect registration of feature point cloud to LiDAR Map (such as associating the feature points of the wall with the ground). Therefore, despite having a very high SR, a wrong registration may lead to subsequent consecutive registration failures, thus triggering more sampling. 
The group intelligence and information sharing mechanism of PSO enable the optimal point to be found quickly, while GRS may easily fall into the local optimal solution. 
PSO and GRS converge quickly and can quickly approach the optimal solution, while AUS requires a large number of sampling points to approach the optimal solution due to its blindness.

\subsection{Evaluation on Dynamic Dataset}
\label{subssec:dynamic}

\begin{figure}[htp]
    \centering
    \includegraphics[width=1.0\linewidth]{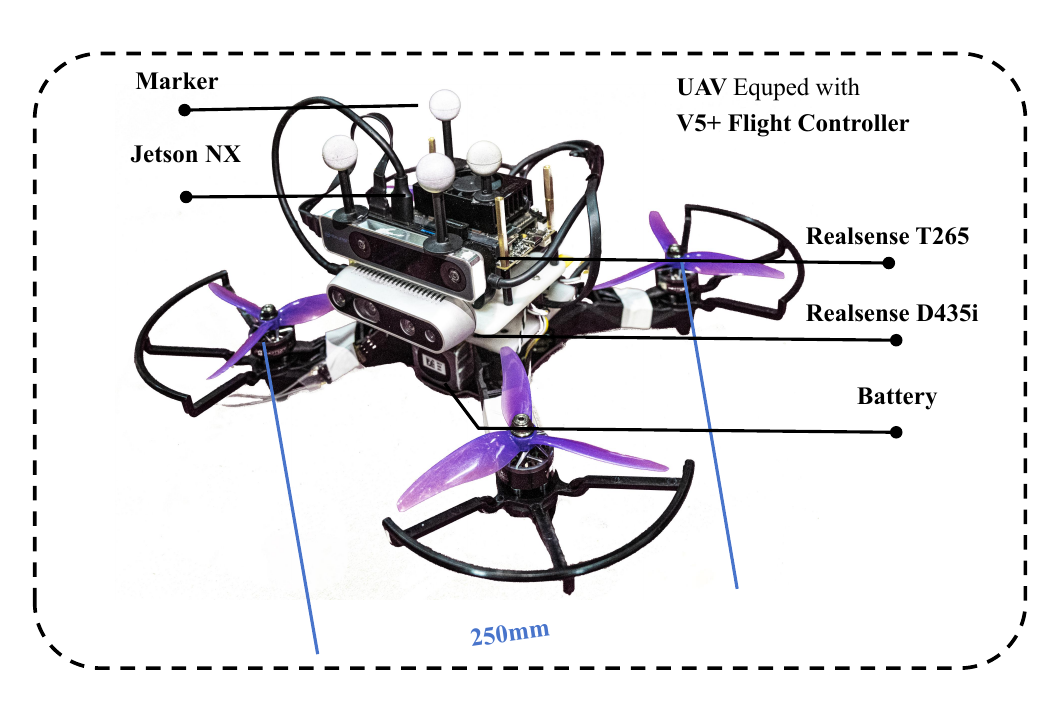}
    \caption{Drone equipped with V5+ flight controller. An Intel RealSense D435i camera is used for collecting synchronized images and IMU data. The onboard Nvidia Jetson NX processes the collected data and transmits it to the UGV in real-time via WiFi. Markers are mounted on drone for visual tracking of motion capture to provide ground truth for the experiment.}
    \label{fig:drone}
\end{figure}

\begin{figure}[htp]
    \centering
    \includegraphics[width=1.0\linewidth]{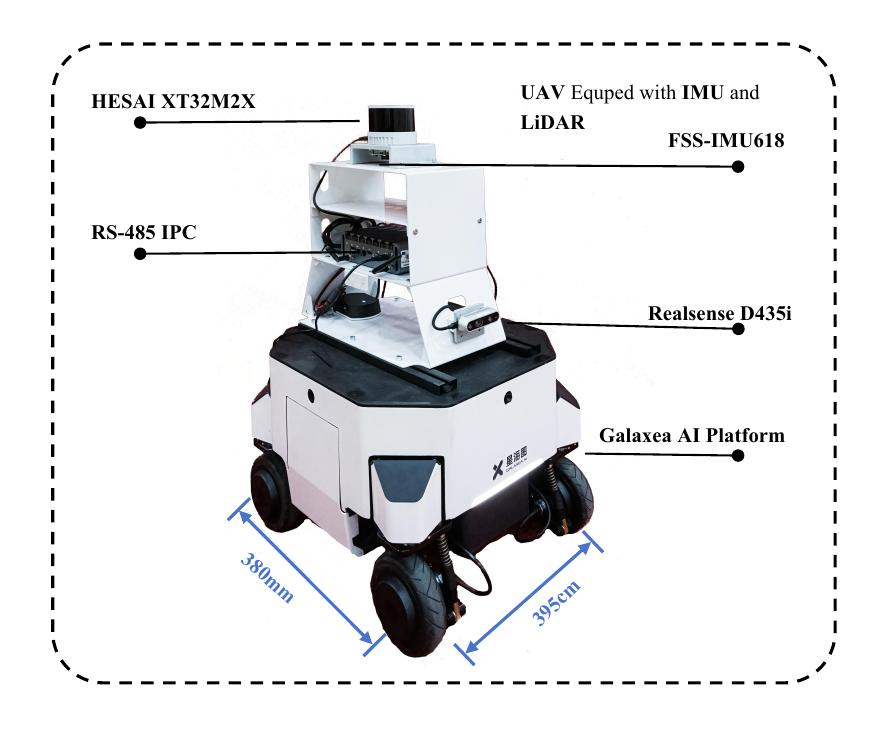}
    \caption{UGV equipped with IMU and LiDAR. The computing platform is an Industrial PC (IPC) with Intel Core i7-4790K CPU, 32GB RAM. Ubuntu 20.04 ROS Noetic is installed on the computer. It collects LiDAR and IMU data from UGVs to build LIO and incremental point cloud map. It also collects images and IMU data from UAVs, and runs the LVCP system in real time.}
    \label{fig:UGV}
\end{figure}

To further evaluate our method in the situation of dynamic point cloud map and multi-agent scenario. We collect real world data by running UGV and UAV at the same time in a room ($35\times 25\times 12m$) equipped with motion capture (FZMotion). The drone is equipped with a Intel RealSense D435i camera to collect synchronized images and IMU data. And we don't use the depth image information of the D435i. The image and imu data are firstly processed by Nvidia Jetson NX and then transmitted to the UGV in real time via WIFI. The UGV carries LiDAR and imu sensor as sensors, and an industrial PC with Intel Core i7-4790K CPU, 32GB RAM as the system’s computing platform. The detailed description of the UAV and UGV platform is shown in Fig. \ref{fig:drone} and Fig. \ref{fig:UGV}.

\begin{figure}
    \centering
    \includegraphics[width=1\linewidth]{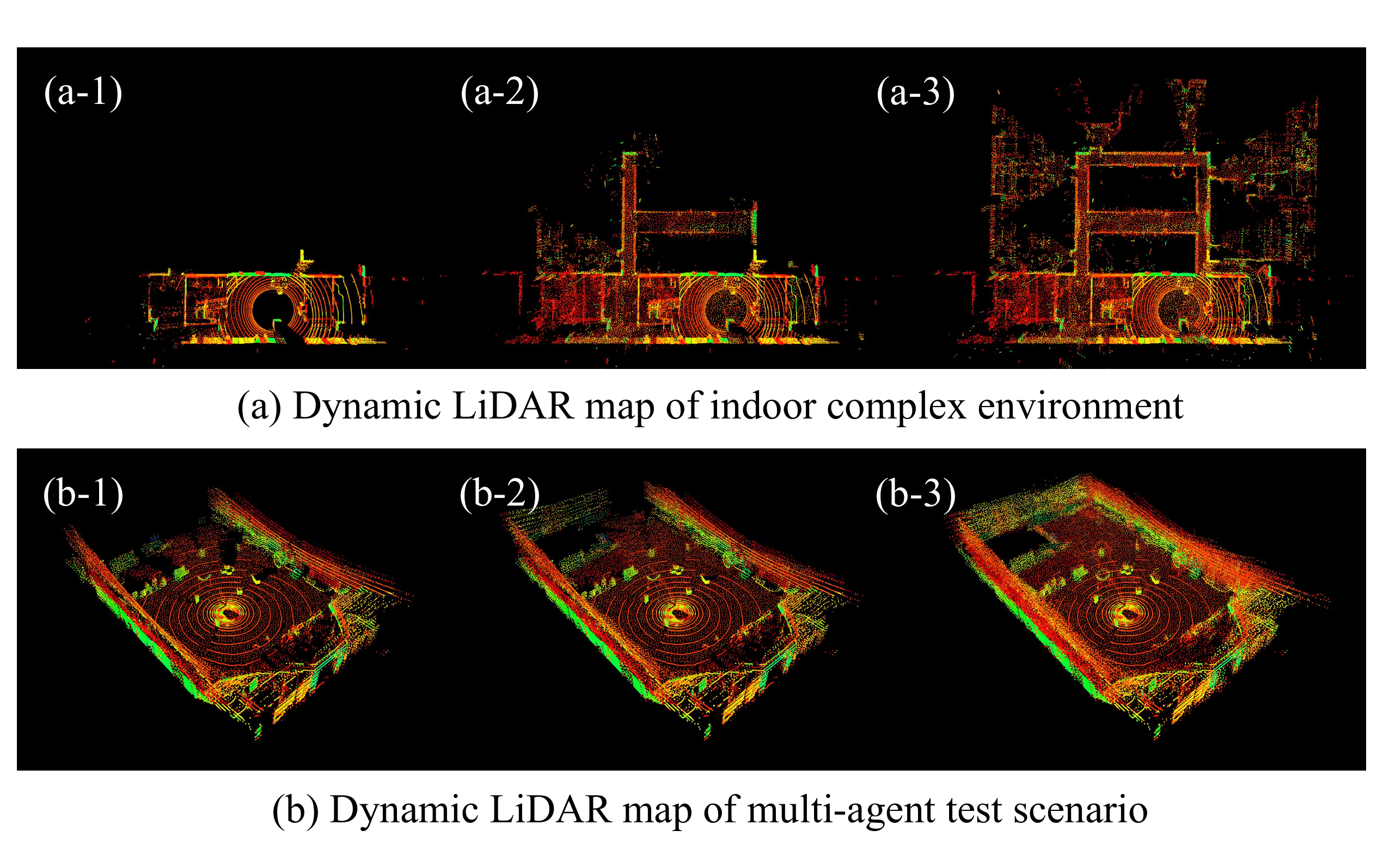}
    \caption{Three snapshots of the dynamic LiDAR point maps in our test scenario}
    \label{fig:dynamic-maps}
\end{figure}

\begin{figure}[htp]
    \centering
    \includegraphics[width=1\linewidth]{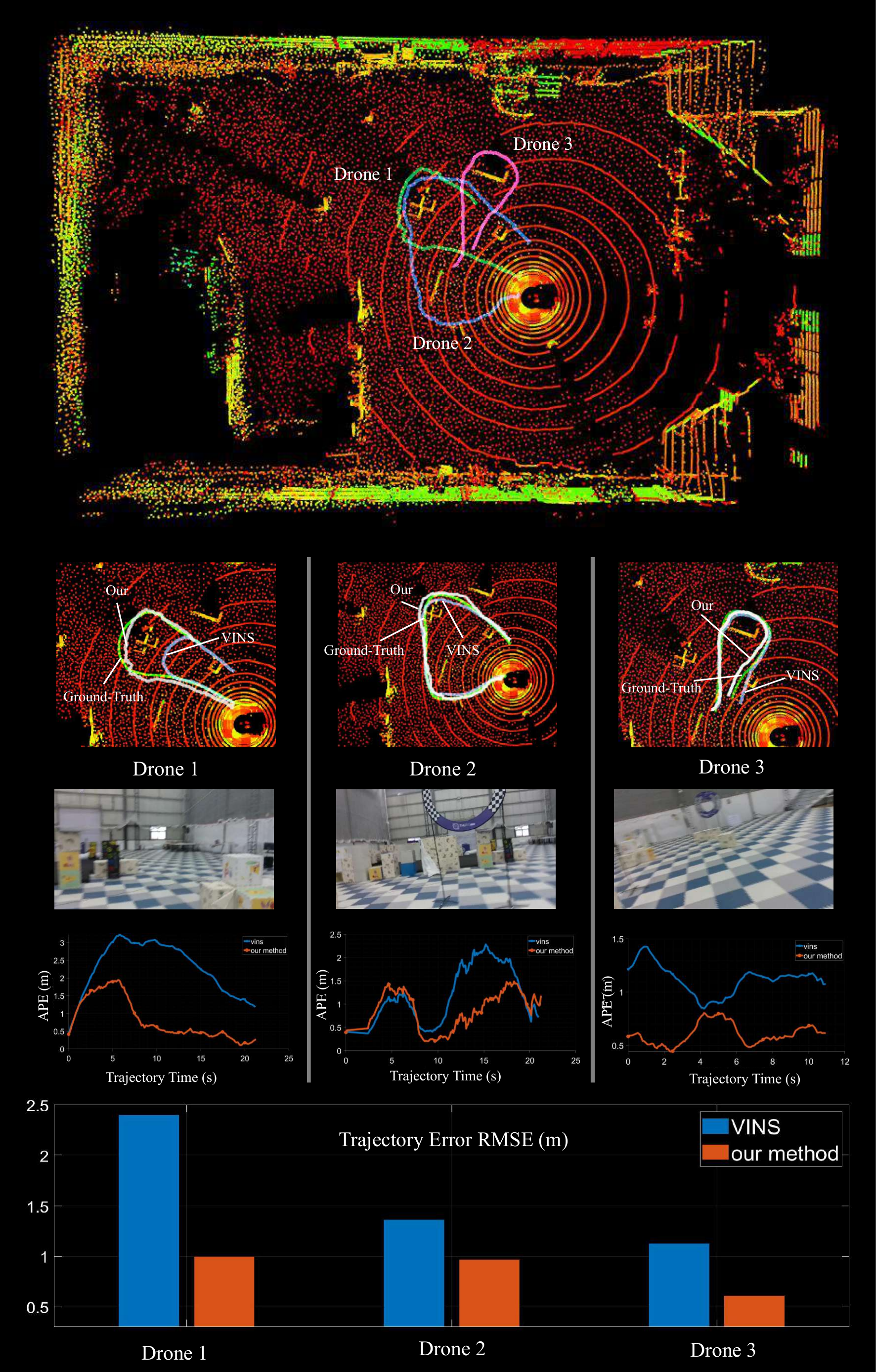}
    \caption{Estimated trajectories in multi-machine scenario. In the upper part of the picture, the trajectories in different colors represent different UAVs, each localizing within a LiDAR map established by the UGV. The initial pose of each UAV is roughly given. They enter the map at different times and their trajectories are estimated. In the lower part of the picture, we compares our method with VINS and ground truth, while the white represents the trajectory estimated by our algorithm, the blue represents the trajectory estimated by VINS, and the green trajectory is the ground truth.}
    \label{fig:changping-multi}
\end{figure}

The dynamic LiDAR point cloud map is incrementally built by LIO, as shown in Fig. \ref{fig:dynamic-maps} (b). In this scenario, multiple drones are controlled by operators to operate in different trajectories. We provide the rough pose of each drone entering the point cloud map, and our initialization module estimates the accurate initial value. Each UAV enter and exit the LiDAR map at different time, which are not known in advance. And the poses of all drones are estimated within a shared LiDAR point cloud map. The estimated trajectorise of the drones are shown in Fig. \ref{fig:changping-multi}.


For each UAV trajectory, we compare our method with VINS and the ground truth. As shown in the lower part of the Fig. \ref{fig:changping-multi}, since VINS does not have the capability to localize within a LiDAR point cloud map, we provid it with a precise initial pose, considering only the trajectory drift and error accumulation due to recursion. 
It can be seen that in situations with significant image jitter and blurriness, monocular VIO algorithms may suffer from severe scale estimation errors and trajectory drift. By using the absolute spatial scale information provided by the LiDAR point clouds for correction, these errors in scale estimation and the accumulation of drift errors can be corrected in real-time. Experiment shows that LVCP system can still work in real time onboard in "$3 \times$UAVs$+$$1 \times$UGV" scenario and ensures effectiveness.


\subsection{Results on More Environments}
\label{subsec:Complex}
In order to verify the generalizability of the LVCP system in the dynamic point cloud scenario, we conduct further experiments in an indoor corridor environment. As shown in Fig.\ref{fig:dynamic-maps} (a), as the UGV moves forward, the LiDAR constructs a 3D structure of the surrounding environment in real time. We hold the RealSense D435i to collect images and IMU data. 




    
\begin{figure}[htp]
    \centering
    \includegraphics[width=1\linewidth]{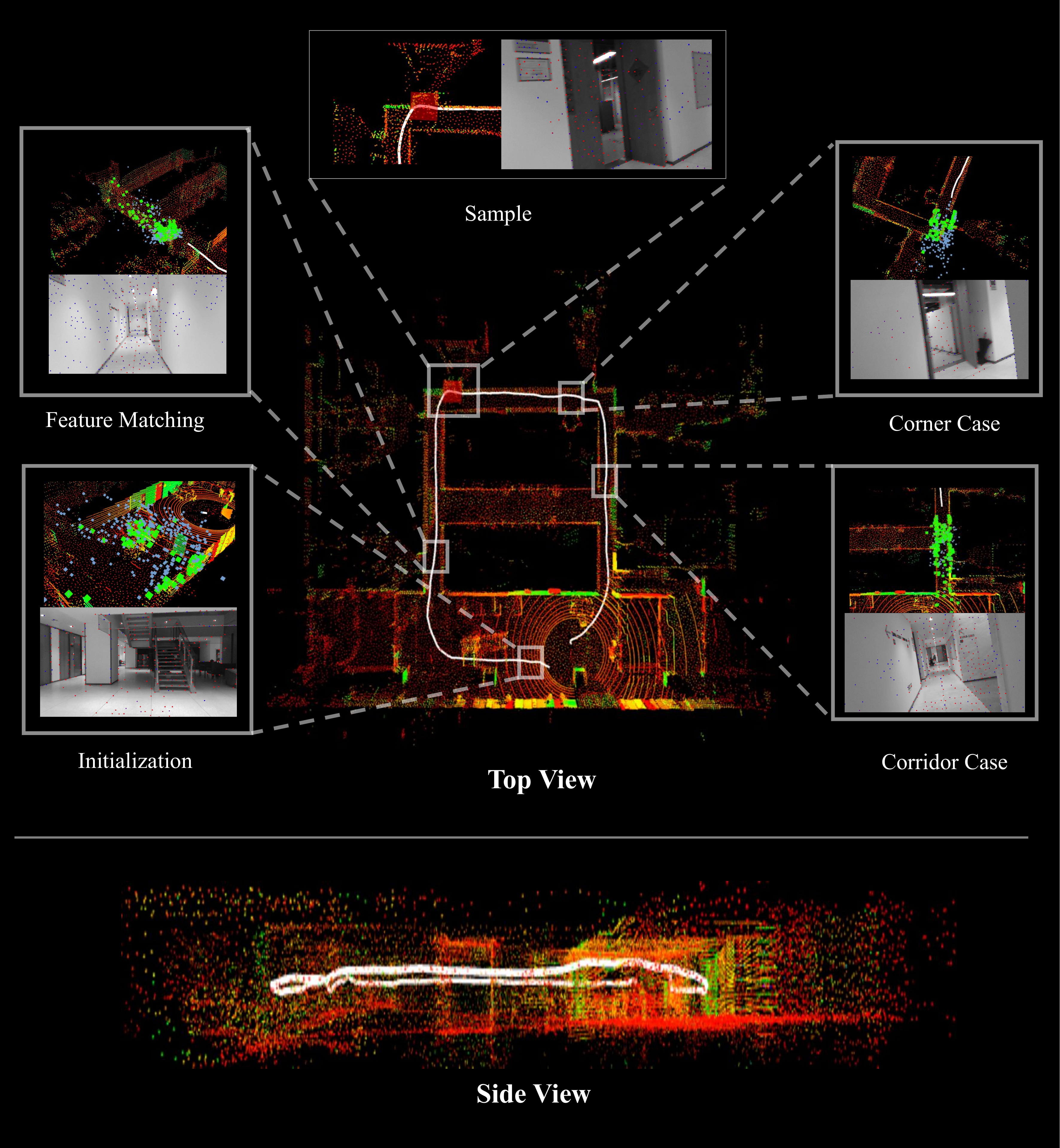}
    \caption{Estimated Trajectory in Indoor Environment. It can be observed that in corridors and at corners, the registration of camera feature point clouds with LiDAR point clouds effectively corrects trajectory drift and estimation errors, constraining the estimated poses within the corridors.}
    \label{fig:traj-xinxilou11}
\end{figure}

Fig.\ref{fig:traj-xinxilou11} describes the camera trajectory estimated by our algorithm within the dynamic indoor corridor point cloud environment. The colored point cloud is acquired by LiDAR and dynamically constructed in real-time based the LIO and current LiDAR scan. The blue point cloud represents the sparse feature point cloud, and the green point could is the matched camera feature point cloud. The white trajectory represents the camera trajectory estimated by our LVCP system. In corridor scenes, it is sometimes difficult for the front end to extract 
enough valid feature points. Due to inaccuracies in initialization and the recursive error inherent in purely VIO approaches, the trajectory estimated by purely VIO tends to deviate at corners. In the LVCP system, feature points are optimized during the Visual-LiDAR Optimization process to accurately match the geometric structure of the surrounding environment. Overall, in the corridor scene, the camera movement is relatively stable, and the PSO sampling is triggered at the corner to correct the camera's pose in time. As the LiDAR moves, the point cloud map becomes denser. In this scenario, the sparsity of the point cloud does not affect the effectiveness of the LVCP system.

\section{Conclusion}
\label{sec:conclusion}

In this paper, we present a robust and real-time method for relative pose estimation between drones and UGV equipped with monocular cameras and LiDAR respectively. The proposed method leverages the complementary strengths of visual and LiDAR data through a tightly coupled framework that does not require prior maps or initial poses. By utilizing LiDAR point clouds to correct VIO drift, our approach ensures accurate and reliable pose estimation even in challenging environments.

Our novel coarse-to-fine framework for LiDAR-visual collaborative localization constructs point-plane associations based on spatial geometric information and employs a new BA problem to estimate the relative pose of the camera and LiDAR simultaneously. The adaptive event-triggered PSO-sampling algorithm enhances the robustness of the system by providing coarse initial pose estimations. Experimental results on both open-source and self-built datasets demonstrate the effectiveness, real-time performance, and robustness of the proposed method, addressing the drift problem in monocular VIO and enabling multiple drone localization tasks.

\section{Discussion}
\label{sec:discuss}

Integrating LiDAR with visual-inertial data for camera pose initialization and real-time pose estimation provides a robust approach for UAV-UGV collaborative exploration and operation. Our approach addresses several key challenges inherent in traditional monocular VIO systems, such as drift over time and the inability to obtain absolute environmental information. By leveraging precise spatial geometry from LiDAR point clouds, we combine sampling and optimization methods to improve the reliability of pose estimation in dynamic and complex environments.

One of the main contributions of this work is the adaptive event-triggered PSO-sampling algorithm, which significantly improves the initialization process by efficiently sampling the pose space and reducing the likelihood of falling into local optima. This method not only accelerates the optimazation process but also ensures that the system can quickly recover from incorrect initial pose estimations, maintaining high accuracy in real-time applications.

However, there are still some limitations and room for future improvement. The quality of the LiDAR point cloud map will have a significant impact on the effect of LVCP system: For areas with sparse point clouds or areas with high noise, there may be incorrect point-plane associations. At this time, the wrong point-to-plane residual in Formula.\ref{equ: point-to-plane residual} will cause the system optimization to go in the wrong direction; in areas not covered by the LiDAR point cloud, the point-to-plane residual in Formula.\ref{equ: point-to-plane residual} will decrease to $0$, and the LVCP system will degenerate into traditional VIO until the camera motion re-observes the LiDAR point cloud or the LiDAR completes the construction of the missing area.

In addition, the LVCP system does not handle dynamic obstacles. The presence of dynamic obstacles will cause some areas in the LiDAR point cloud map to change frequently, resulting in map instability and inconsistency, making it difficult for the system to maintain an accurate representation of the environment. And it is a challenge for front-end feature matching and point-to-plane association.


In conclusion, our tightly coupled LiDAR-visual collaborative positioning framework represents a significant step forward in achieving robust and real-time relative pose estimation for air-ground collaborative tasks. This work lays a strong foundation for future research and development in the field, with the potential to enable more advanced and reliable autonomous systems.

\appendix 
\subsection{Derivation of Point-to-plane Residual}
\label{subsec:plane}

Given a plane, its plane equation is described as:
\begin{equation}
\label{equ:plane}
      \begin{aligned} 
      Ax+By+Cz+D=0
   \end{aligned} 
\end{equation}
The equation can be transformed into:
\begin{equation}
      \begin{aligned} 
      \frac{A}{D}x+\frac{B}{D}y+\frac{C}{D}z=-1
   \end{aligned} 
\end{equation}
Then only three variables are needed to determine a plane. Use QR to solve the overdetermined equation:
\begin{equation}
      \begin{aligned} 
      \left[ \begin{matrix}
	x_1&		y_1&		z_1\\
	x_2&		y_2&		z_2\\
	x_3&		y_3&		z_3\\
	x_4&		y_4&		z_4\\
	x_5&		y_5&		z_5\\
\end{matrix} \right] \left[ \begin{array}{c}
	\frac{A}{D}\\
	\frac{B}{D}\\
	\frac{C}{D}\\
\end{array} \right] =-1
   \end{aligned} 
\end{equation}
For a more intuitive representation, $[\frac{A}{D},\frac{B}{D},\frac{C}{D}]^T$ can be written as a normalized vector $n =\left[ a,b,c \right] ^T$ and its modular inverse $w$. 

For each point in space $\left[ x_0,y_0,z_0 \right] ^T \in\mathbb{R} ^3$ , let the foot of the perpendicular be $\left[ x_1,y_1,z_1 \right] ^T \in\mathbb{R} ^3$. The vector from $\left[ x_0,y_0,z_0 \right]$ to the foot of the perpendicular $\left[ x_1,y_1,z_1 \right]$ is $\vec{d}=\left[ x_1-x_0,y_1-y_0,z_1-z_0 \right] ^T$. Since $\vec{d}$ is perpendicular to the plane, we have:
\begin{equation}
\label{equ:abc2}
      \begin{aligned} 
d\left[ \begin{array}{c}
	a\\
	b\\
	c\\
\end{array} \right] =\left[ \begin{array}{c}
	x_1-x_0\\
	y_1-y_0\\
	z_1-z_0\\
\end{array} \right] 
   \end{aligned} 
\end{equation} 
or
\begin{equation}
\label{equ:-abc}
      \begin{aligned} 
d\left[ \begin{array}{c}
	a\\
	b\\
	c\\
\end{array} \right] =-\left[ \begin{array}{c}
	x_1-x_0\\
	y_1-y_0\\
	z_1-z_0\\
\end{array} \right] 
   \end{aligned} 
\end{equation} 
Put point $\left[ x_1,y_1,z_1 \right]$ into the equation \ref{equ:plane} and \ref{equ:abc2}, we can obtain:
\begin{equation}
      \begin{aligned} 
a\left( da+x_0 \right) +b\left( db+y_0 \right) +c\left( dc+z_0 \right) +1=0
   \end{aligned} 
\end{equation}
so point-to-plane distance $d$ can be obtained:

\begin{equation}
\label{equ:residual}
      \begin{aligned} 
d=\left| ax_0+by_0+cz_0+w \right|   
\end{aligned} 
\end{equation}
Put point $\left[ x_1,y_1,z_1 \right]$ into the equation \ref{equ:plane} and \ref{equ:-abc}, we can get the same result.

\subsection{Derivation of IMU-integration}
\label{subsec:IMU}

The IMU pre-integration technique was first proposed in \cite{lupton2011visual} and was subsequently extended to Lie algebra in a series of works \cite{forster2015manifold}, forming an elegant and comprehensive theoretical system. This theory has now been widely applied in VIO and LIO based on the Bundle Adjustment optimization framework, becoming the fundamental algorithm for handling IMU data within optimization frameworks, including a series of excellent works such as ORBSLAM3\cite{campos2021orb}, VINS\cite{qin2018vins}, and LIO-SAM\cite{liosam2020shan}.

The raw measurements from the accelerometer and gyroscope are given by:
\begin{equation}
\begin{aligned}
\hat{\boldsymbol a}_t &= \boldsymbol a_t + \boldsymbol b_a + R_t^w g + n_a, \\
\hat{\boldsymbol \omega}_t &= \boldsymbol \omega_t + \boldsymbol b_\omega + n_\omega
\end{aligned}
\end{equation}
where $\hat{\boldsymbol a}_t$ and $\hat{\boldsymbol \omega}_t$ are the measured linear acceleration and angular velocity $\boldsymbol a_t$ and $\boldsymbol \omega_t$ are the true linear acceleration and angular velocity $\boldsymbol b_a$ and $\boldsymbol b_\omega$ are the accelerometer and gyroscope biases $R_t^w$ is the rotation matrix from the world frame to the IMU frame $g$ is the gravity vector,$n_a$ and $n_\omega$ are the measurement noise terms.

To derive Equ.\ref{equ:IMU factor}, we start by integrating the IMU measurements to propagate the state between two consecutive frames. This integration process is essential to reduce the computational load and avoid repeated integration during optimization. The derivation involves transforming the IMU measurements from the world frame to the local frame and pre-integrating them.

The state propagation between two image time instants, corresponding to IMU frames $I_k$ and $I_{k+1}$, can be described using the raw gyroscope $\hat{\boldsymbol a}_t$ and accelerometer $\hat{\boldsymbol {\omega}}_t$ measurements, respectively:

\begin{equation}
\begin{aligned}
&
p^w_{I_{k+1}} = p^w_{I_k} + v^w_{I_k} \Delta t_k \\&
+ \int_{t \in [t_k, t_{k+1}]} (R^w_t (\hat{\boldsymbol a}_t -\boldsymbol b_{a_t} - n_a) - g_w) dt^2, \\&
v^w_{I_{k+1}} = v^w_{I_k} + \int_{t \in [t_k, t_{k+1}]} (R^w_t (\hat{\boldsymbol a}_t -\boldsymbol b_{a_t} - n_a) - g_w) dt, \\&
q^w_{I_{k+1}} = q^w_{I_k} \otimes \int_{t \in [t_k, t_{k+1}]} \frac{1}{2} \Omega(\hat{\boldsymbol \omega}_t - \boldsymbol b_{\omega_t} - n_w) q^k_t dt,
\end{aligned}
\end{equation}
where $\Omega (\omega)$ is the angular velocity matrix defined as:
\begin{equation}
\begin{aligned}
\Omega (\omega )=\left[ \begin{matrix}
	-|\omega |_{\times}&		\omega\\
	-\omega ^T&		0\\
\end{matrix} \right] 
\end{aligned}
\end{equation}
and $\Omega (\omega)$ is the skew-symmetric matrix of the angular velocity $\omega$. 

To reduce the computational cost, we pre-integrate the IMU measurements relative to the local frame $I_k$. This approach allows us to pre-integrate the parts related to linear acceleration $\hat{\boldsymbol a}_t$ and angular velocity $\hat{\boldsymbol \omega}_t$:

\begin{equation}
\begin{aligned}
&
R_{I_{k}}^w p^w_{I_{k+1}} = R_{I_{k}}^w (p^w_{I_{k}} + v^w_{I_{k}} \Delta t_k \\&
- \frac{1}{2} g_w \Delta t_k^2) + \alpha^{I_k}_{I_{k+1}}, \\&
R_{I_k}^w v^w_{I_{k+1}} = R_{I_K}^w (v^w_{I_k} - g_w \Delta t_k) + \beta^{I_k}_{I_{k+1}}, \\&
q_{I_k}^w \otimes q^w_{I_{k+1}} = \gamma^{I_k}_{I_{k+1}},
\end{aligned}
\end{equation}
where the pre-integration terms are defined as:
\begin{equation}
\begin{aligned}
&
\alpha^{I_k}_{I_{k+1}} = \int_{t \in [t_k, t_{k+1}]} R_{I_k}^t (R^w_t (\hat{\boldsymbol a}_t -\boldsymbol b_{a_t} - n_a) - g_w) dt^2, \\&
\beta^{I_k}_{I_{k+1}} = \int_{t \in [t_k, t_{k+1}]} R_{I_k}^t (R^w_t (\hat{\boldsymbol a}_t -\boldsymbol b_{a_t} - n_a) - g_w) dt, \\&
\gamma^{I_k}_{I_{k+1}} = \int_{t \in [t_k, t_{k+1}]} \frac{1}{2} \Omega(\hat{\boldsymbol \omega}_t - \boldsymbol b_{\omega_t} - n_w) \gamma_{I_k}^t dt.
\end{aligned}
\end{equation}
These pre-integration terms $\alpha^{I_k}_{I_{k+1}} $, $\beta^{I_k}_{I_{k+1}}$ and $\gamma^{I_k}_{I_{k+1}}$ are computed based on the IMU measurements within the time interval $[t_k, t_{k+1}]$ with $I_k$ as the reference frame. This strategy saves significant computational resources during optimization because we do not need to re-propagate IMU measurements repeatedly.



{
    \bibliographystyle{IEEEtran}
    \bibliography{IEEEabrv, bibliography}
}

\end{document}